\documentclass[lettersize,journal]{IEEEtran}
\usepackage{amsmath,amsfonts}
\usepackage{algorithmic}
\usepackage[ruled,vlined,linesnumbered]{algorithm2e}
\usepackage{array}
\usepackage[caption=false,font=normalsize,labelfont=sf,textfont=sf]{subfig}
\usepackage{textcomp}
\usepackage{stfloats}
\usepackage{url}
\usepackage{verbatim}
\usepackage{graphicx}
\usepackage{cite}
\usepackage{balance}
\usepackage{paralist}
\usepackage{booktabs}
\usepackage{hyperref}
\usepackage{xspace}
\usepackage{color}
\usepackage{multirow}
\usepackage{ifthen}

\renewcommand{\baselinestretch}{1}

\long\def\ignorethis#1{}

\definecolor{gray}{rgb}{0.35,0.35,0.35}
\definecolor{MyBlue}{rgb}{0,0.2,0.8}
\definecolor{MyRed}{rgb}{0.8,0.2,0}
\definecolor{MyGreen}{rgb}{0.0,0.5,0.1}
\definecolor{MyGray}{rgb}{0.4,0.4,0.4}

\def\red#1{\textcolor{red}{#1}}
\def\blue#1{\textcolor{blue}{#1}}
\def\green#1{\textcolor{green}{#1}}

\newlength\paramargin
\newlength\figmargin
\newlength\subfigmargin
\newlength\secmargin
\newlength\subsecmargin
\newlength\tabmargin
\newlength\eqmargin

\usepackage{array}
\newcolumntype{L}[1]{>{\raggedright\let\newline\\\arraybackslash\hspace{0pt}}m{#1}}
\newcolumntype{C}[1]{>{\centering\let\newline\\\arraybackslash\hspace{0pt}}m{#1}}
\newcolumntype{R}[1]{>{\raggedleft\let\newline\\\arraybackslash\hspace{0pt}}m{#1}}

\def\ie{i.e.,~}
\def\eg{e.g.,~}
\def\etc{etc}

\newcommand{\secref}[1]{Section~\ref{sec:#1}}

\newcommand{\figref}[1]{Fig.~\ref{fig:#1}}
\newcommand{\tabref}[1]{Table~\ref{tab:#1}}
\newcommand{\eqnref}[1]{Eq.\eqref{Eq:#1}}

\newcommand{\algref}[1]{Algorithm~\ref{#1}}

\newcommand{\Paragraph}[1]{\noindent\textbf{#1}}

\usepackage{capt-of}
\usepackage{color}
\usepackage{colortbl}
\usepackage{tabularx} 
\definecolor{sp}{RGB}{31,78,121}
\definecolor{sy}{RGB}{127,96,0}
\definecolor{mycolor_blue}{RGB}{231,239,250}
\definecolor{mycolor_green}{RGB}{230,247,224}
\definecolor{mycolor_gray}{RGB}{236,236,236}
\definecolor{pearDark!20}{RGB}{212,230,241}
\usepackage{listings}
\usepackage{xcolor}
\usepackage{float}
\usepackage{cleveref}
\definecolor{codegreen}{rgb}{0,0.6,0}
\definecolor{codegray}{rgb}{0.5,0.5,0.5}
\definecolor{codepurple}{rgb}{0.58,0,0.82}
\definecolor{backcolour}{rgb}{0.95,0.95,0.92}

\lstdefinestyle{mystyle}{
    backgroundcolor=\color{backcolour},   
    commentstyle=\color{codegreen},
    keywordstyle=\color{magenta},
    numberstyle=\tiny\color{codegray},
    stringstyle=\color{codepurple},
    basicstyle=\ttfamily\footnotesize,
    breakatwhitespace=false,         
    breaklines=true,                 
    captionpos=b,                    
    keepspaces=true,                 
    numbers=left,                    
    numbersep=5pt,                  
    showspaces=false,                
    showstringspaces=false,
    showtabs=false,                  
    tabsize=2
}

\newfloat{lstfloat}{htbp}{lop}
\floatname{lstfloat}{Algorithm}
\crefalias{lstfloat}{listing}

\lstdefinestyle{myverbatim}{
    basicstyle=\ttfamily\footnotesize,
    backgroundcolor=\color{white},
    breaklines=true,
    breakatwhitespace=true
}

\begin{document}
\title{StarVid: Enhancing Semantic Alignment in Video Diffusion Models 
\\
via \underline{S}patial and Syn\underline{T}actic Guided \underline{A}ttention \underline{R}efocusing}
\author{Yuanhang Li, Qi Mao, Lan Chen, Zhen Fang, Lei Tian, Xinyan Xiao, Libiao Jin, Hua Wu
\thanks{$^ \ast$Corresponding author: Qi Mao (e-mail: qimao@cuc.edu.cn).
}}

\markboth{Journal of \LaTeX\ Class Files,~Vol.~14, No.~8, August~2021}%
{Shell \MakeLowercase{\textit{et al.}}: A Sample Article Using IEEEtran.cls for IEEE Journals}


\maketitle

\begin{abstract}

Recent advances in text-to-video (T2V) generation with diffusion models have garnered significant attention. 
However, they typically perform well in scenes with a single object and motion, struggling in compositional scenarios with multiple objects and distinct motions to accurately reflect the semantic content of text prompts.
To address these challenges, we propose \textbf{StarVid}, a plug-and-play, training-free method that improves semantic alignment between multiple subjects, their motions, and text prompts in T2V models. 
StarVid first leverages the spatial reasoning capabilities of large language models (LLMs) for two-stage motion trajectory planning based on text prompts.
Such trajectories serve as spatial priors, guiding a spatial-aware loss to refocus cross-attention (CA) maps into distinctive regions.
Furthermore, we propose a syntax-guided contrastive constraint to strengthen the correlation between the CA maps of verbs and their corresponding nouns, enhancing motion-subject binding.
Both qualitative and quantitative evaluations demonstrate that the proposed framework significantly outperforms baseline methods, delivering videos of higher quality with improved semantic consistency.
\end{abstract}


\section{Introduction}
\label{sec:intro}

\begin{figure*}[!t]
    \centering
     \includegraphics[width=1\linewidth]{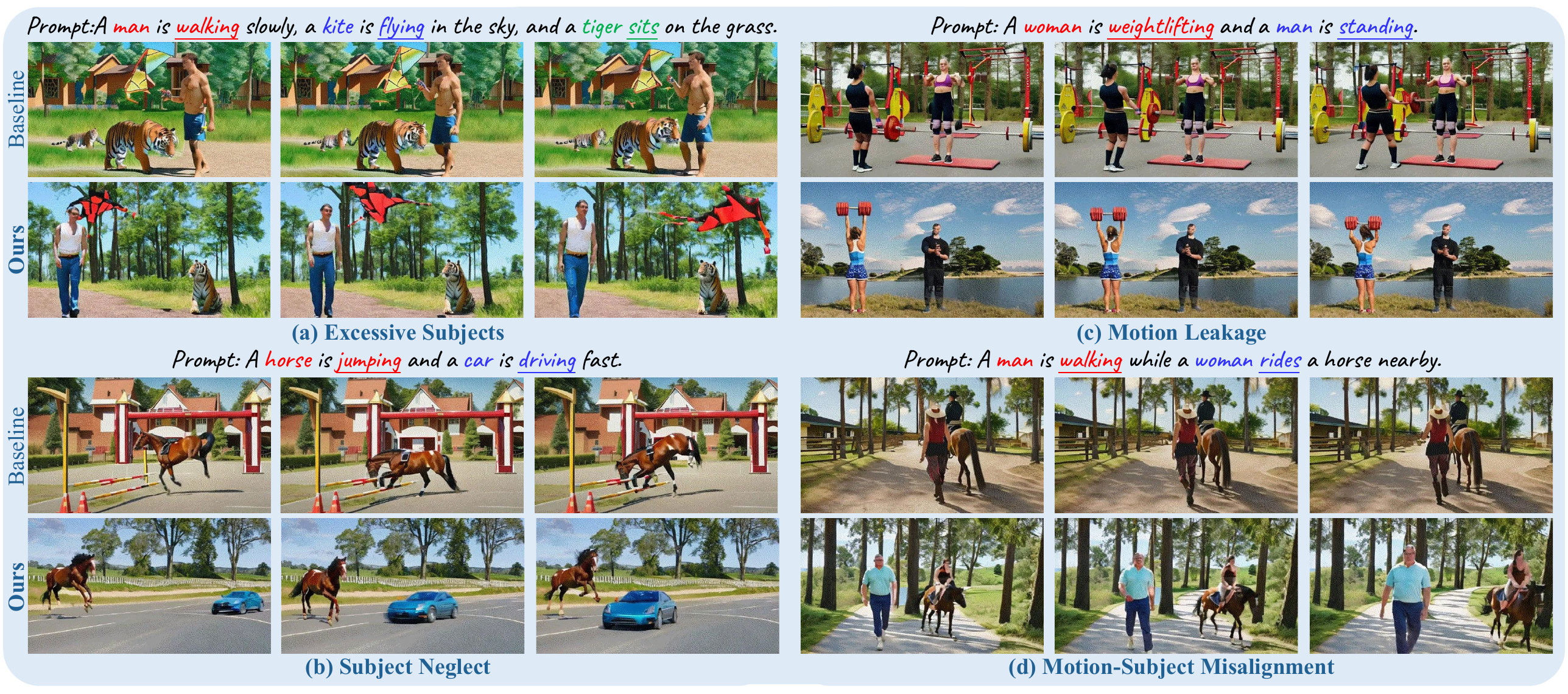}
     \vspace{-7 mm}
      \caption{
    \textbf{Semantic misalignment in T2V diffusion models.}
    In compositional scenarios involving {\textbf{multiple subjects}} with {\textbf{distinctive motions}}, the generated videos often fail to align accurately with textual descriptions, leading to discrepancies in subject count and issues with incorrect motion binding.
    Our method, \textbf{StarVid}, a training-free approach, harnesses the capabilities of LLMs and incorporates both spatial and syntax-aware attention-refocusing guidance to improve the semantic alignment of multiple subjects and their respective motions.
    }
    \vspace{-5 mm}
    \label{fig:teaser}
\end{figure*}

%
%
%
\IEEEPARstart{I}{n} recent years, 
significant progress has been made in diffusion-based text-to-image (T2I) generation models~\cite{10528891,10589534,10021829,rombach2022high,10480591,ramesh2022hierarchical,saharia2022photorealistic,10261222,10812849}, enabling the creation of visually high-quality images that correspond closely to the provided text prompts.
Building on this achievement, researchers have expanded the scope of diffusion models to include text-to-video (T2V) generation~\cite{wang2023modelscope,chen2024videocrafter2,khachatryan2023text2video,yu2024efficient,pikalab,Dreamina}.
%
Although several models based on U-Net~\cite{ronneberger2015u} or DiT~\cite{peebles2023scalable} architectures are capable of generating high-quality videos that align with textual semantics, their performance tends to be more effective in relatively straightforward scenarios featuring a \textbf{\emph{single}} dominant object with a \textbf{\emph{single}} motion.
%
However, when users express the need for more compositional scenarios involving \textbf{\emph{multiple}} objects with \textbf{\emph{distinct}} motions, existing models may encounter difficulties in accurately reflecting the semantics of text prompts.
%
%
%
As summarized in \figref{teaser}, semantic misalignment in these models include 
a) \textbf{\emph{subject count mismatch}}, where these models may generate excessive subjects (\figref{teaser} (a)) or subject neglect (\figref{teaser}(b)), leading to inconsistent motion correspondence in the video;
b) \textbf{\emph{incorrect motion binding}}, where even when the correct number of subjects is identified, associating the motion with its corresponding subject remains challenging, resulting in motion leakage (\figref{teaser}(c)) or motion-subject misalignment (\figref{teaser}(d)).
Recent studies~\cite{chefer2023attend, kim2023dense, phung2023grounded, rassin2024linguistic} have addressed semantic misalignment in text-to-image (T2I) models, particularly for multiple-object compositional generation.
A primary issue identified is numerical inconsistency, which T2V models often inherit when initialized with the spatial components of pre-trained T2I models. 
Additionally, the absence of a direct connection between text descriptions and the temporal module in T2V models further complicates the alignment of objects with their corresponding motions.
Despite these challenges, there remains a notable gap in the literature concerning the enhancement of semantic alignment in T2V models.

To bridge this gap, we first unveil that the cross-attention (CA) maps of nouns and verbs in U-Net-based T2V diffusion models~\cite{wang2023modelscope} can effectively capture the spatial layout and trajectory of motions, respectively.
Subsequently, 
we make two key observations towards CA maps of both nouns and verbs on video-text misalignment and alignment examples using ZeroScope~\cite{wang2023modelscope}: 
%
First, the CA maps corresponding to subject nouns fail to converge within defined areas in the early denoising timesteps, lacking clear differentiation from one another. 
%
This, in turn, obstructs the concentration of high-attention areas of verbs into specific regions. 
%
Second, the CA maps associated with verbs struggle to accurately pinpoint the regions where their corresponding subjects are situated, thereby leading to issues such as motion leakage and inbinding.
%
Consequently, an effective solution requires the \textbf{\emph{1) distinctly localization of the CA maps for nouns and 2) the subsequent alignment of these maps with the CA maps of their corresponding verbs.}}

%
%
Building on the observations discussed, 
we introduce \textbf{StarVid}, a novel training-free, plug-and-play approach that refocuses CA maps in T2V models to better align subjects, their motions, and the semantics of the text prompts.
In particular, we begin by harnessing the spatial reasoning capabilities of Large Language Models (LLMs)~\cite{achiam2023gpt} to parse text prompts incrementally through a two-stage motion trajectory planner. 
This planner generates spatial layout trajectories that adhere to subject numeracy and physical principles.
Subsequently, these motion trajectories serve as spatial layout guidance, with specifically designed special-aware CA-based constraints ensuring that the CA maps of nouns and verbs distinctly localize to specific regions.

However, we observe that the spatial guidance alone cannot fully prevent the CA maps of the verbs from attending to other regions, resulting in motion leakage and misalignment issues.
To address this, we propose a syntax-guided contrastive constraint to minimize the distance between the CA maps of verbs and their corresponding nouns relative to other words, thereby strengthening the association between subjects and their motions.
Moreover, we introduce a multi-frame strategy for constructing positive and negative pairs, thus ensuring consistent motion across different frames.
\figref{teaser} demonstrates the effectiveness of the proposed method compared to T2V baselines, significantly enhancing semantic alignment in terms of both subject numeracy correctness and motion binding.

Our contributions can be summarized as follows:

\begin{compactitem}
\item We propose a plug-and-play, training-free method designed to enhance the semantic alignment within existing T2V models when text prompts involve multiple objects with distinct motions.

\item We investigate integrating LLMs with a two-stage motion trajectory planner to generate spatial layouts from text prompts, thereby directing spatial and syntactic attention-refocusing constraints to accurately position subjects in appropriate regions and establish connections with their motions.

\item Extensive quantitative and qualitative experiments demonstrate the effectiveness of our proposed method against other baselines using the benchmark with multiple subjects and diverse motions.
\end{compactitem}

\begin{figure*}[!t]
    \centering
    \includegraphics[width=1\linewidth]{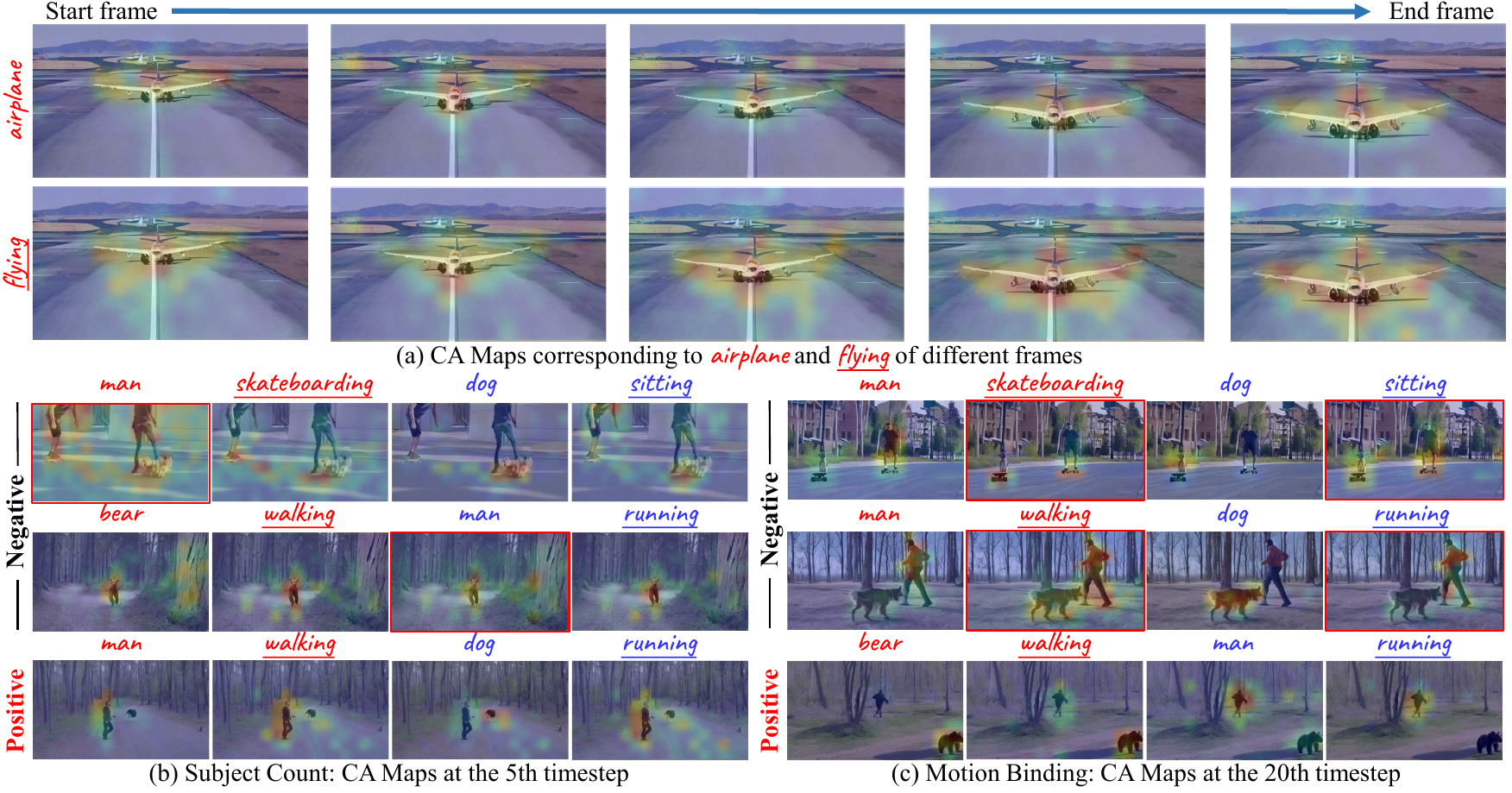}   
    \vspace{-8 mm}
    \caption{\textbf{Visualization of CA maps of nouns and verbs using ZeroScope~\cite{wang2023modelscope}.}
     (a)  The spatial trajectory of ``\emph{\red{\underline{flying}}}'' is effectively captured by the CA maps and correlates with shifts in the CA maps of the noun ``\emph{\red{airplane}}''.
     (b)
         During visualization, we examine examples of video-text misalignment (denoted as negative) and alignment (denoted as positive). In the early stage of denoising, we observe that the high-attention areas of nouns either overlap or are globally dispersed, resulting in instances of subject neglect and subject increase.
     (c) At a later denoising stage, the CA map of the verb fails to align with the CA map of the noun, leading to motion leakage and incorrect binding.
     }
     \vspace{-5 mm}
    \label{fig:fig_2}
\end{figure*}

\section{Related work}
\label{sec:related_work}

\subsection{Text-to-Video Diffusion Models}
%
%
%
%
%
Diffusion-based T2V generation models~\cite{wang2023modelscope, singer2022make, yuan2023instructvideo, yu2024efficient} have made significant progress in recent years. 
%
Several existing studies~\cite{khachatryan2023text2video, yuan2023instructvideo, guo2023animatediff} have focused primarily on improving the temporal consistency of generated videos.
For example, InstructVideo~\cite{yuan2023instructvideo} enhances models using human feedback, whereas AnimateDiff~\cite{guo2023animatediff} maintains fixed pre-training weights and updates only the motion modeling module.
%
Additionally, some studies ~\cite{zhang2023controlvideo, lian2023llmgroundedvideo, lin2023videodirectorgpt, yang2024direct, yin2023dragnuwa} introduce additional conditions such as depth map ~\cite{zhang2023controlvideo}, bounding box ~\cite{lian2023llmgroundedvideo,lin2023videodirectorgpt, yang2024direct}, and motion trajectory ~\cite{yin2023dragnuwa} to control the shape and motion of the subject.
%
%
%
However, most existing research~\cite{lian2023llmgroundedvideo, yang2024direct} concentrates on scenarios featuring a single object performing a single motion.
%
In scenes with multiple subjects performing various motions, the issue of semantic misalignment becomes prominent yet underexplored. 

\subsection{Attention Refocusing}
%
%
%
Attention-refocusing techniques have been extensively explored in T2I generation~\cite{chefer2023attend,chen2024training,xie2023boxdiff,phung2023grounded, rassin2024linguistic}.
Some studies~\cite{chefer2023attend,rassin2024linguistic} focus on developing attention-based constraints to address the issue of prompt unfollowing in T2I generation.
Attend-and-Excite~\cite{chefer2023attend} proposes a loss function designed to directly modulate the attention weights assigned to nouns, thereby addressing the issue of subject neglect in image generation.
SynGen~\cite{rassin2024linguistic} employs the linguistic structure within prompts to tackle the issues of subject neglect and attribute leakage in image generation.
Other research~\cite{xie2023boxdiff, chen2024training} introduces additional spatial priors to improve subject positioning.
Box-diff~\cite{xie2023boxdiff} employs bounding boxes to modulate the attention maps of the subject.
Additionally, some works~\cite{couairon2023zero, mao2023mag} utilize masks to confine the CA map's region and reduce attribute leakage in image generation and editing. 
%
In the field of T2V generation, recent advances~\cite{lian2023llmgroundedvideo, chen2024motion}, also incorporate spatial priors with attention-based guidance to control object motions. 
However, the issue of semantic misalignment when dealing with multiple subject-motion scenarios remains underexplored, which is the primary focus of this paper.

\subsection{LLM-Assisted Compositional Generation}
LLMs enhance vision generation models through their powerful reasoning capabilities. 
Several studies~\cite{NEURIPS2023_13250eb1, phung2023grounded, lian2023llm, feng2024layoutgpt, zhang2024realcompo} have harnessed these capabilities to improve the alignment between the output images of 
T2I generation models and their associated text prompts.
Additionally, other research~\cite{lian2023llmgroundedvideo, lin2023videodirectorgpt, lu2023flowzero} utilizes text prompts fed into LLMs to generate dynamic layouts, which in turn guide the motion trajectories of subjects in generated videos. 
However, directly generating these dynamic layouts places a significant planning burden on LLMs, often leading to outcomes that conflict with real-world physical laws.
To address this, our paper proposes a two-stage approach, employing LLMs to strategically reduce their cognitive load.
%

%
\section{Methodologies}
%
%
%
%

In this section, we first discuss the rationale for enhancing semantic correspondence through CA map refocusing, as detailed in \secref{motivation-1}. 
%
We then analyze the underlying causes of \textbf{\emph{subject count mismatch}} and \textbf{\emph{incorrect motion binding}} by visualizing the CA maps of nouns and verbs in \secref{motivation-2}.
%
This analysis leads to two key observations that motivate us to develop StarVid (\secref{our_method}) to improve multiple subject-motion correspondence.


\begin{figure*}[!t]
    \centering
    \includegraphics[width=1\linewidth]{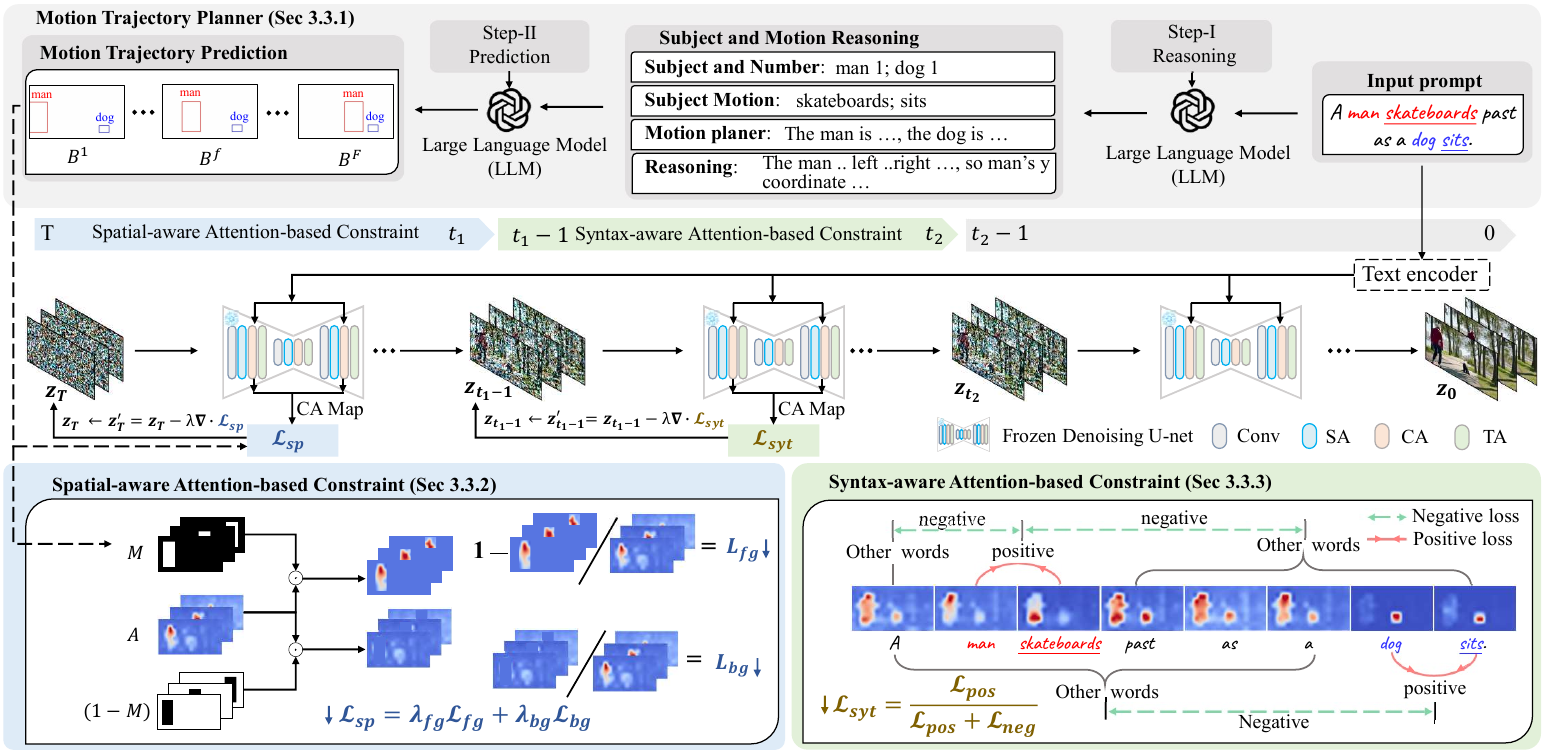}    
    \vspace{-7 mm}
    \caption{
    \textbf{Overview of our StarVid.}
    Given a prompt (\eg ``\emph{A \red{man} \red{\underline{skateboards}} past as a \blue{dog} \blue{\underline{sits}}}.''), we first employ a two-stage motion trajectory planner LLM to progressively plan the motion trajectories of the subjects (\eg ``\red{man}'', ``\red{dog}''), ensuring alignment with real-world dynamics.
   In the early denoising step, we construct a spatial-aware attention-based constraint \textcolor{sp}{ $\mathcal{L}_{sp}$ } that guides the CA maps of nouns (\eg ``\red{man}'') and verbs (\eg ``\red{\underline{skateboards}}'') to specific spatial locations. 
   In the subsequent denoising phase, we introduce a syntax-aware attention-based constraint \textcolor{sy}{$\mathcal{L}_{syt}$} that reduces the distance between the CA maps of a verb (\eg ``\red{\underline{skateboards}}'') and its corresponding noun (\eg ``\red{man}''), while increasing separation from other words. 
    }
    \vspace{-5 mm}
    \label{fig:framework}
\end{figure*}

\subsection{Why Adjust the CA Map?}
\label{sec:motivation-1}
%
%
%

Let a T2V model generate a video $v \in F\times H\times W \times C$, where $H$, $W$, $F$, and $C$ indicate the height, width, number of frames, and number of channels, respectively.
The backbone diffusion models typically employ a U-Net architecture, with a core module that integrates the input text information and the frame features through the cross-attention (CA) layer.
In the CA layer, we denote the correlation between visual features and the $i$-th word in the $f$-th frame as $A_{i}^f \in \mathbb{R}^{h\times w}$, where $h \ll H$ and $w \ll W$ are the height and width of the visual feature map.
By visualizing the CA map, we can observe the area where the text prompt influences frame $f$.
As demonstrated in \figref{fig_2}(a), the spatial trajectory of ``\emph{\red{\underline{flying}}}'' is effectively represented by the CA map and correlates with shifts in the CA map of the noun ``\emph{\red{airplane}}''.
Consequently, by controlling the attention regions of the CA Maps for nouns and verbs, we can influence the object's motion trajectory and establish its connection to the motion.

\subsection{Motivations}
\label{sec:motivation-2}

As discussed in \secref{motivation-1}, the CA maps of verbs roughly capture the spatial trajectory of the motion.
%
To better understand the increasing occurrences of \textbf{\emph{subject count mismatch}} and \textbf{\emph{incorrect motion binding}}, we first conduct experiments using ZeroScope~\cite{wang2023modelscope}. 
The experiments utilize text descriptions based on the template: ``a \red{object1} is \red{\underline{motion1}} and a \blue{object2} is \blue{\underline{motion2}}''.
%
%
%
Next, we compare the CA Maps of semantic alignment (positive examples) and misalignment (negative examples) across different denoising timesteps.

\textbf{Observation 1:} 
\textit{
The CA maps of nouns related to different objects are separate and focus on distinctive regions, which ensures the convergence of subjects into correct numbers.}
The early stage of the denoising process in the T2V generation model usually determines the overall layout of the objects.
As shown in \figref{fig_2}(b, first two rows), at the $5$-th denoising timestep, the highly correlated regions within the nouns' CA maps do not rapidly converge and separate effectively.
As a result, the number of subjects fails to align with the text description.
As illustrated in \figref{fig_2}(b, last row), the high-attention regions on the nouns' CA maps are relatively focused and distinctly separated, leading to the accurate generation of the correct number of subjects.

\textbf{Observation 2:} 
\textit{
Ensuring strong connections between CA maps of verbs and nouns helps enhance motion correspondence.
}
As illustrated in \figref{fig_2}(c, first two rows), by the $20$-th denoising timestep, although the high-attention areas on the CA maps for nouns align with the corresponding objects, the CA maps for verbs still fail to accurately target the correct objects. 
%
This leads to motion leakage and incorrect binding. 
%
One potential reason is the use of the CLIP~\cite{radford2021learning} text encoder, known for its inability to effectively encode linguistic structures~\cite{feng2022training}, resulting in a lack of clear relationship between verbs and nouns, and thus misalignment in the CA maps between verbs and their corresponding objects. 
%
However, as demonstrated in \figref{fig_2}(c, last row), the CA maps for the verbs focus solely on the appropriate subject areas, ensuring that the subjects' motion aligns with the text description.
%

\subsection{Our Solution: StarVid}
\label{sec:our_method}

In this paper, we aim to enhance the semantic alignment in multiple subjects with distinctive motions for existing T2V models.
The proposed StarVid pipeline is illustrated in \figref{framework}. 
Given a text prompt $P$ consisting of $L$ words, we define the set of words in $P$ as $S= \{ s_{1}, s_{2}, \cdots, s_{L} \}$.
We also define the set of noun and verb pairs in $P$ as $S^\ast $,  $s_i^\ast =  \{s_i, s_j \} \in S^\ast$ (where $s_i$ is a noun, $s_j$ is a verb, $i$, and $j$ are respective indices in $S$).
We first utilize the LLM model, \ie GPT-4o~\cite{gpt4o} to analyze the nouns and verbs in $P$, which automatically generates each subject's motion trajectory for subsequent spatial guidance.
To address issues of \textbf{\emph{subject count mismatch}} and \textbf{\emph{incorrect motion binding}}, we propose spatial-aware and syntax-aware constraints, denoted as \textcolor{sp}{$\mathcal{L}_{sp}$} and \textcolor{sy}{$\mathcal{L}_{syt}$}, respectively.

\subsubsection{Using LLM as Motion Trajectory Planner}
\label{sec:llm}
%

%
%
%
Inspired by real-world filmmaking, where a director first determines the number of actors and planned actions, and then orchestrates their behaviors and movements before shooting the scene, we adopt a similar methodology in our model. 
Instead of directly using an LLM~\cite{achiam2023gpt} to predict subjects' motion trajectories~\cite{lian2023llmgroundedvideo}, we introduce a Chain-of-Thought (CoT)~\cite{wei2022chain} strategy. 
This strategy entails designing a two-stage motion planner that comprises the following two components: 

%
\begin{compactitem}
\item \textbf{Subject and Motion Reasoning.}
Given a text prompt $P$, the LLM first predicts explicit information such as the subject, the number of subjects, and their motions. 
Additionally, it performs simple motion planning and explains the reasoning behind its decisions.
\item \textbf{Motion Trajectory Prediction.}
After subject and motion reasoning, the text prompt $P$ and explicit information are fed to the LLM again to predict the subject's dynamic motion trajectory $B=\{B_i^f \}$, where $B_i^f$ denotes the layout of the $f$-th frame corresponding to the $i$-th subject.
Each $B_i^f$ contains top-left and bottom-right coordinates. 
\end{compactitem}
%
%

\begin{algorithm}
\caption{A Denoising Step Using StarVid}
\label{algorithm}
\SetAlgoLined
\KwIn{A text prompt $P$; a set od noun and verb pairs $S^{\ast}$; a set of spatial masks $M$ derived by LLM; a timestep $t$ and the noise features $z_t$; the timestep $t_1$ and $t_2$; the maximum iteration step $iter_1$ and $iter_2$; the hyperparamters $\alpha $, $\lambda _1$, $\lambda _2$; a function $F_{1}(\cdot)$ and a function $F_{2}(\cdot)$ for computing the proposed constraint \textcolor{sp}{$\mathcal{L}_{\rm{sp}}$} and \textcolor{sy}{$\mathcal{L}_{\rm{syt}}$}; a pre-trained Video Diffusion model $VD$.}
\KwOut{The noise latent $z_{t-1}$ for the next timestep.}
\If{$t \le t_1$}{
    \For{$i = 1$ \KwTo $iter_1$}{
        $\_, A_t \gets VD(z_t,\mathcal{P},t)$ \;
        $\mathcal{L}_{\rm{sp}} \gets F_{1}(A_t, M, S^{\ast})$ \;
        ${z}' _t \gets z_t - \alpha \lambda _1 \mathcal{L}_{\rm{sp}}$\;
        $z_t \gets {z}'_t$ \;
    }
}
\If{$t_1 < t \le t_2$}{
    \For{$i = 1$ \KwTo $iter_2$}{
        $\_, A_t \gets VD(z_t,\mathcal{P},t)$ \;
        $\mathcal{L}_{\rm{syt}} \gets F_{2}(A_t, M, S^{\ast})$ \;
        ${z}' _t \gets z_t - \alpha \lambda _2 \mathcal{L}_{\rm{syt}}$\;
         $z_t \gets {z}'_t$ \;
    }
}
$z_{t-1} \gets VD(z_t,P,t)$ \;
\Return{$z_{t-1}$}\;
\end{algorithm}

%
%
\subsubsection{Injecting Motion Trajectory as Spatial Prior}
\label{sec:spatial}
Based on the \textbf{Observation 1}, our initial goal is to adjust the CA maps of nouns during the early denoising stages, ensuring that their attention areas become concentrated and distinctly separated from each other.
To achieve this, we utilize dynamic motion trajectory $B$, generated by \secref{llm}, to guide nouns quickly to focus on specified regions.
A set of spatial masks $M=\{M_i^f\}$ is obtained by transforming $B$, where the value inside the box is $1$ and the value outside the box is $0$.
To ensure that the CA maps of nouns concentrate on regions defined by the given spatial prior, we propose a \textcolor{sp}{\textbf{spatial-aware constraint}} aimed at enhancing the focus of these CA maps on the foreground objects,
\vspace{-1 mm}
\begin{equation}
   \mathcal{L}_{\rm{fg}}=\frac{1}{F} \sum_{i,j\in S^\ast} \sum_{f \in F}  \left ( 1-  \frac {A_i^f \cdot M_i^f }{A_i^f} \right )^2.
   \label{Eq:L_Fg}
\vspace{-1 mm}
\end{equation}

Focusing exclusively on the information within the motion trajectory ensures that the subject remains within the specified range; however, it does not prevent nouns from considering information outside the bounding boxes, potentially leading to the generation of multiple subjects outside the designated region.
Therefore, we propose an additional background constraint aimed at minimizing the influence of the CA maps of nouns outside the bounding boxes, as follows,
\vspace{-1 mm}
\begin{equation}
   \mathcal{L}_{\rm{bg}}= \frac{1}{F} \sum_{i,j \in S^\ast} \sum_{f \in F}\left ( \frac {A_i^f \cdot \left ( 1 - M_i^f \right )}{A_i^f} \right )^2.
   \label{Eq:L_bg}
   \vspace{-1 mm}
\end{equation}

Consequently, the overall spatial-aware attention-based constraint can be formulated as,
\vspace{-1 mm}
\begin{equation}
   \textcolor{sp}{\mathcal{L}_{\rm{sp}}}=\lambda _{fg} \mathcal{L}_{fg} + \lambda _{bg} \mathcal{L}_{bg}.
\label{Eq:Lspatialguidance}
\vspace{-1 mm}
\end{equation}

Moreover, considering that the CA maps of verbs $A_j^f$ should align with those of nouns for motion correspondence, we also apply \eqnref{Lspatialguidance} to the CA maps of verbs to reinforce this alignment using the same bounding boxes with the corresponding nouns.

\begin{figure}[!t]
    \centering
    \includegraphics[width= 0.9\linewidth]{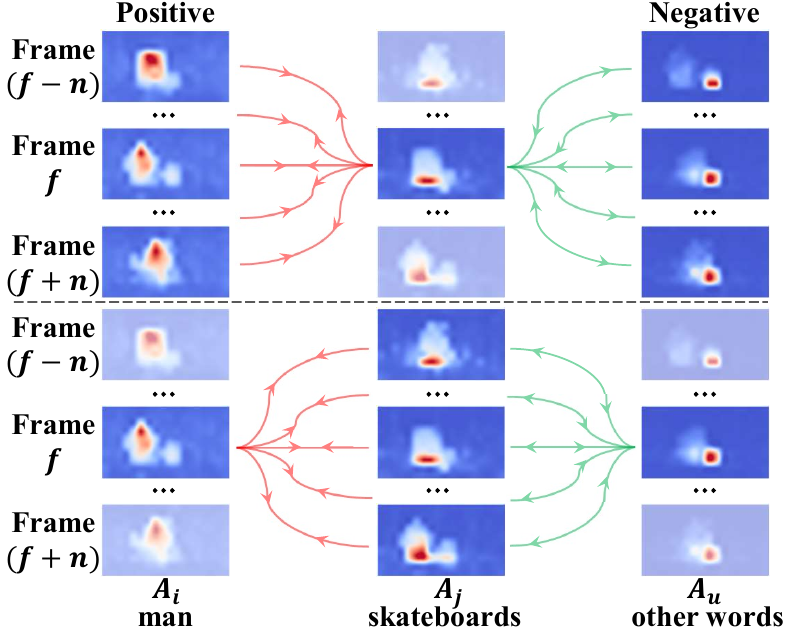}    
    \caption{
    \textbf{Illustration of our multi-frame contrastive strategy.}
    The CA Map of the verb for the $f$-th frame should be closer to the CA Maps of the noun and farther from those of other words in the $f$-th and adjacent frames, and vice versa.
    }
    \vspace{-5 mm}
    \label{fig:multi_frame_pipeline}
\end{figure}

\subsubsection{Enhancing Motion and Subject Correspondence}
\label{sec:camapContLoss}
Under the spatial constraints applied to both nouns and verbs, the CA maps of verbs can align to some extent with the regions defined by the spatial priors. 
However, a clear relationship between the CA maps of verbs and nouns is still lacking, resulting in the leakage of verb CA maps.
According to \textbf{Observation 2}, our goal is to leverage the syntactic relationships to establish a strong connection between verbs and nouns, thereby enhancing the alignment between motion and corresponding subjects. 
As a result, our approach involves introducing contrastive learning to minimize the distance between the CA map of verbs and that of the corresponding nouns, while simultaneously distancing it from the CA maps of other words to prevent interference.

In particular, we construct a \textbf{\textcolor{sy}{syntax-aware contrastive constraint}}, where the noun and verb pairs in each $s_i^ \ast$ serve as positive samples for each other, while other words in $S$ act as their negative samples.
Additionally, to ensure motion consistency across adjacent frames, as illustrated in \figref{multi_frame_pipeline}, we propose a \textbf{multi-frame contrastive strategy} that incorporates these adjacent frames into the calculation of the 
$f$-th frame, thereby effectively expanding the contrastive space.
For the noun $s_i$ and the verb $s_j$ in $s_i^ \ast$, our positive loss aims to minimize the distance between the CA map of $s_i$ and $s_j$,

\begin{equation}
{\begin{split}
\vspace{-3 mm}
   \mathcal{L}_{pos} ( s_i^ \ast) = \frac{1}{F} \sum_{f \in F} \frac{1}{4n+1}\left (\sum_{k}^{[f \pm n]}  f_{dist}\left ( A_{i}^{f}, A_{j}^{k} \right ) \right. \\ \left.
+ \sum_{k, k\ne f}^{[f \pm n]}  f_{dist}\left ( A_{i}^{k}, A_{j}^{f} \right ) \right ),
   \label{Eq:L_pos}
    \vspace{-3 mm}
  \end{split}}
\end{equation}
where $f_{dist}(\cdot)$ represents the calculation of the distance function between CA maps, and $n$ represents the number of adjacent frames of the $f$-frame.
For $s_i^\ast$, we define the set of other words in $S$ as $U_i$.
Our negative loss encourages the separation of $i$-th word pairs from the words in $U_i$,

\vspace{-2 mm}
{\begin{equation}
\begin{split}
  \mathcal{L}_{neg} ( s_i^ \ast, U_i) =\frac{1}{F} \sum_{u \in U_i} \sum_{f \in F} \frac{1}{4n+1}\left( \sum_{k}^{[f \pm n]} f_{dist}\left ( A_i^{f}, A_u^{k} \right ) \right. \\
 \left. \sum_{k, k\ne f}^{[f \pm n]} f_{dist}\left ( A_i^{k}, A_u^{f} \right ) \right ) .
   \label{Eq:L_neg}
   \vspace{-2 mm}
\end{split}
\end{equation}}

Finally, the syntax-aware attention-based constraint can be formulated as follows,
\begin{equation}
   \textcolor{sy}{\mathcal{L}_{\rm{syt}}} = \sum_{s_i^\ast \in S^\ast} \frac{\mathcal{L}_{pos} ( s_i^ \ast)}{\mathcal{L}_{pos} ( s_i^ \ast) + \mathcal{L}_{neg} ( s_i^ \ast, U_i)}.
   \label{Eq:L_camapc}
\end{equation}

\subsubsection{Attention Refocusing via Latent Optimization}

After obtaining the constraints, we compute their gradients to update the noise latent $z_t$ at each timestep as follows,
\begin{equation}
   {z}'_{t} \longleftarrow z_{t}-\alpha \cdot \lambda_{*} \nabla L_{*},  
\label{Eq:L_optim}
\end{equation}
where $\alpha$ represents the learning rate of the optimization process, and $\lambda_{*}$ controls the weighting of the constraint.
Specifically, we initially apply \eqnref{Lspatialguidance} in the first $t_1$ steps out of $50$ denoising steps.
The spatial-aware guidance first optimizes $z_t$ to gradually align with the motion trajectory.
This process accurately generates the correct number of subjects and ensuring that their motions correspond to the positions of the related objects.
Subsequently, \eqnref{L_camapc} is implemented over the next $t_2 - t_1$ steps to further refine the $z_t$ shift, enhancing the high-response attention alignment between nouns and verbs.
The implementation details refer to \algref{algorithm}.
\section{Experiments}

\begin{table*}[!t]
\centering
\caption{\textbf{Automatic evaluations results of Action Binding and LLM-Generated benchmark.}
The best values for ZeroScope~\cite{wang2023modelscope} and VideoCrafter2~\cite{chen2024videocrafter2} are highlighted in \colorbox{pearDark!20}{blue} and \colorbox{mycolor_green}{green}, respectively.
}
\resizebox{\linewidth}{!}{%
\begin{tabular}{cccccc|ccccc}
\hline
\multicolumn{1}{c}{} & \multicolumn{5}{c}{\textbf{Action Binding Benchmark}} &\multicolumn{5}{c}{\textbf{LLM-Generated Benchmark}} \\ \cmidrule(r){2-6} \cmidrule(r){7-11} 
\multicolumn{1}{c}{} & \multicolumn{2}{c}{\textbf{Video Quality}} & \multicolumn{3}{c}{\textbf{Semantic Alignment}} & \multicolumn{2}{c}{\textbf{Video Quality}} & \multicolumn{3}{c}{\textbf{Semantic Alignment}} \\ \cmidrule(r){2-3} \cmidrule(r){4-6} \cmidrule(r){7-8} \cmidrule(r){9-11}
\multicolumn{1}{c}{\multirow{-4}{*}{\textbf{Method/Metrics}}} & \textbf{Pick Score} \textbf{($\uparrow$)} & \textbf{CLIP-I} \textbf{($\uparrow$)} & \textbf{CLIP-T} \textbf{($\uparrow$)} & \textbf{Numeracy} \textbf{($\uparrow$)} & \textbf{Action Binding} \textbf{($\uparrow$)}& 
\textbf{Pick Score} \textbf{($\uparrow$)} & \textbf{CLIP-I} \textbf{($\uparrow$)} & \textbf{CLIP-T} \textbf{($\uparrow$)} & \textbf{Numeracy} \textbf{($\uparrow$)} & \textbf{Action Binding} \textbf{($\uparrow$)} \\ 
\midrule
 ZeroScope~\cite{wang2023modelscope} & 20.67 & \colorbox{pearDark!20}{0.97} & 26.98 & 0.435 & 0.551 & 20.54 & \colorbox{pearDark!20}{0.94} & 26.44 & 0.536 & 0.646 \\
 LVD~\cite{lian2023llmgroundedvideo} & 19.91 & 0.96 & 25.47 & 0.645 & 0.571 & 20.21 & 0.92 & 25.67 & 0.716 & 0.647 \\
 DAV~\cite{yang2024direct} & 19.85 & 0.94 & 21.08 & 0.379 & 0.434 & 19.39 & 0.93 & 22.75 & 0.503 & 0.420 \\
 \textbf{Ours} & \colorbox{pearDark!20}{20.73} & \colorbox{pearDark!20}{0.97} & \colorbox{pearDark!20}{28.02} & \colorbox{pearDark!20}{0.678} & \colorbox{pearDark!20}{0.674} & \colorbox{pearDark!20}{20.69} & \colorbox{pearDark!20}{0.94} & \colorbox{pearDark!20}{27.76} & \colorbox{pearDark!20}{0.871} & \colorbox{pearDark!20}{0.795} \\ 
 \midrule
 VideoCrafter2~\cite{chen2024videocrafter2} & 21.24 & \colorbox{mycolor_green}{0.97} & 27.12 & 0.543 & 0.531 & 21.17 & \colorbox{mycolor_green}{0.96} & 27.16 & 0.694 & 0.632 \\
\textbf{Ours} & \colorbox{mycolor_green}{21.49} & \colorbox{mycolor_green}{0.97} & \colorbox{mycolor_green}{28.96} & \colorbox{mycolor_green}{0.713} & \colorbox{mycolor_green}{0.724} & \colorbox{mycolor_green}{21.47} & \colorbox{mycolor_green}{0.96} & \colorbox{mycolor_green}{28.74} & \colorbox{mycolor_green}{0.908} & \colorbox{mycolor_green}{0.809} \\ 
\bottomrule
\end{tabular}%
}
\label{tab:tab_1}
\end{table*}

\begin{figure*}[!t]
    \centering
    \includegraphics[width=1\linewidth]{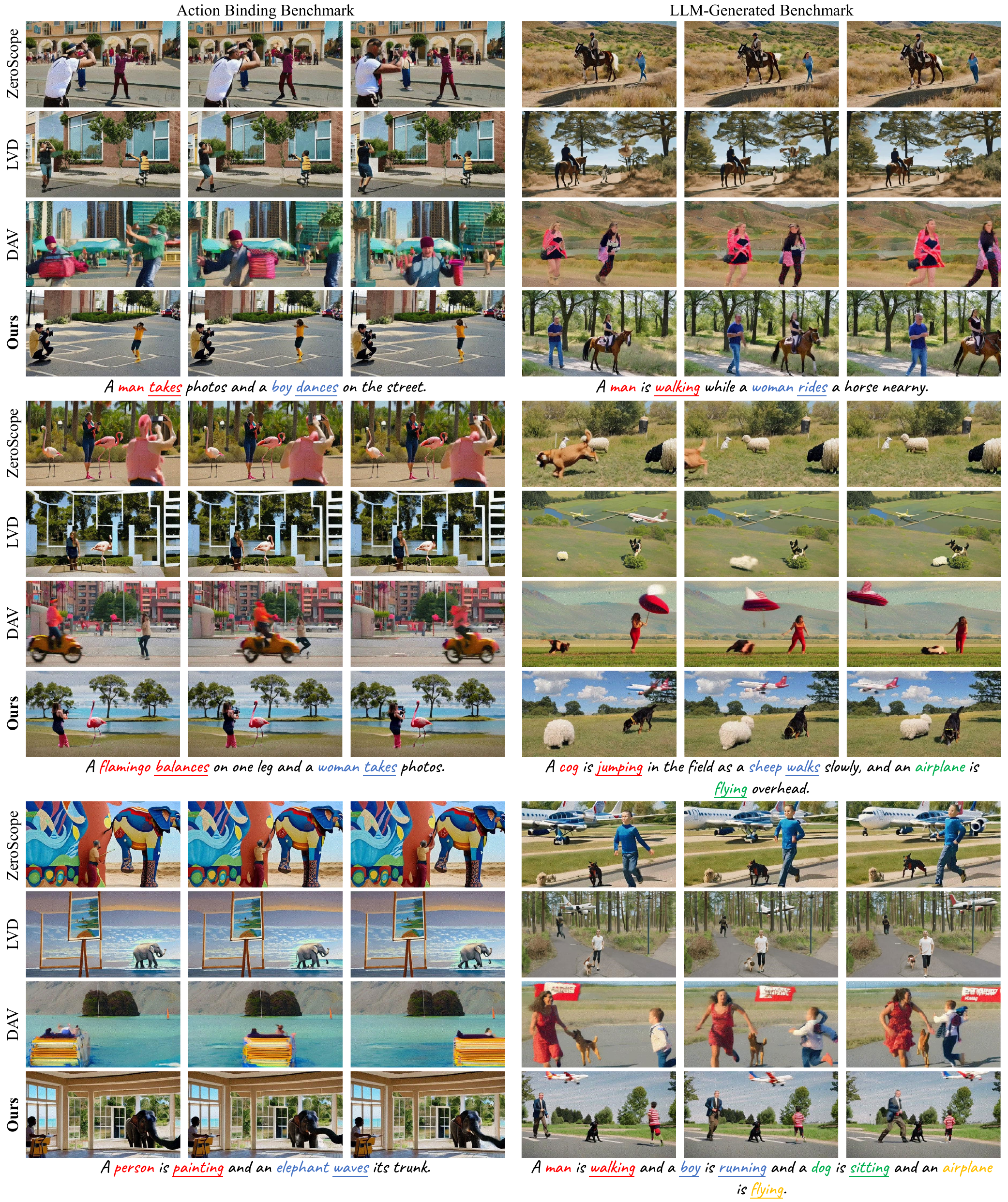}
    \vspace{-7 mm}
    \caption{\textbf{Qualitative comparisons with ZeroScope~\cite{wang2023modelscope}.}
    Our proposed \textbf{StarVid} not only accurately generates the requisite number of subjects but also effectively associates them with their respective motions. 
    }
    \vspace{-5 mm}
    \label{fig:fig_4}
\end{figure*}

\subsection{Experiment Setup}

\Paragraph{Implementation Details.} 
%
We adopt the open-source T2V generative model ZeroScope~\cite{wang2023modelscope} and VideoCrafter2~\cite{chen2024videocrafter2} as our backbone and apply our method on top of it.
%
All generated videos are $16$-frame sequences. 
For the baseline models, the resolution of ZeroScope~\cite{wang2023modelscope} is set to $320 \times 576$, and the resolution of VideoCrafter2~\cite{chen2024videocrafter2} is set to $320 \times 512$.
Our experiments are carried out on a single V100 GPU.
We set $t_1$ and $t_2$ to $5$ and $25$, respectively.
%
We use the DDIM scheduler~\cite{song2020denoising} to denoise each generation over $50$ timesteps.
%
For $\textcolor{sp}{\mathcal{L}_{\rm{sp}}}$ guidance, we apply it only to the initial $5$ timesteps, with a maximum of $10$ iterations per timestep.
%
In \eqnref{Lspatialguidance}, the weights for both $\lambda _{fg}$ and $\lambda _{bg}$ are set to $1$. 
%
In \eqnref{L_optim}, the loss weight $\lambda_{*}$ for $\textcolor{sp}{\mathcal{L}_{\rm{sp}}}$ is $30$, and the learning rate $\alpha$ is $1$. 
%
From the $6$-th to the $25$-th timestep, $\textcolor{sy}{\mathcal{L}_{\rm{syt}}}$ applies once at each timestep to align the subjects and their motions.
%
We use the Kullback-Leibler (KL) divergence to calculate the distance between CA Maps, with the number of adjacent frames set to $1$, in formulas \eqnref{L_pos} and \eqnref{L_neg}. 
%
In \eqnref{L_optim}, the loss weight  $\lambda_{*}$ for $\textcolor{sy}{\mathcal{L}_{\rm{syt}}}$ is $20$, and the learning rate $\alpha$ is $1$.

\Paragraph{Benchmarks.} 
To validate the effectiveness of our method StarVid, and to facilitate comparisons with other methodologies, we employ the following two benchmarks:
\begin{compactitem}
\item \textbf{Action Binding Benchmark.} 
This benchmark comes from T2V-CompBench~\cite{sun2024t2v} and evaluates the ability of T2V generation models to associate actions with their corresponding objects.
It includes 100 text prompts, each featuring two objects and their corresponding actions, generated by LLM~\cite{achiam2023gpt}.
\item \textbf{LLM-Generated Benchmark.} 
To verify the ability of our model and other models involving more than two subjects and their respective motions, we collect $26$ subjects and $18$ motions.
Then, we employ the LLM~\cite{achiam2023gpt} to mimic human language patterns, automatically generating text prompts based on subjects and motions.
We generate $200$ prompts for this benchmark.
Each text prompt contains \textbf{two or more} subjects along with their respective motions, such as ``\emph{A \red{man} is \red{\underline{walking}} slowly, a \blue{kite} is \blue{\underline{flying}} in the sky, and a \green{tiger} \green{\underline{sits}} on the grass.}''
\end{compactitem}
%

\begin{figure*}[h]
    \centering
    \includegraphics[width=1\linewidth]{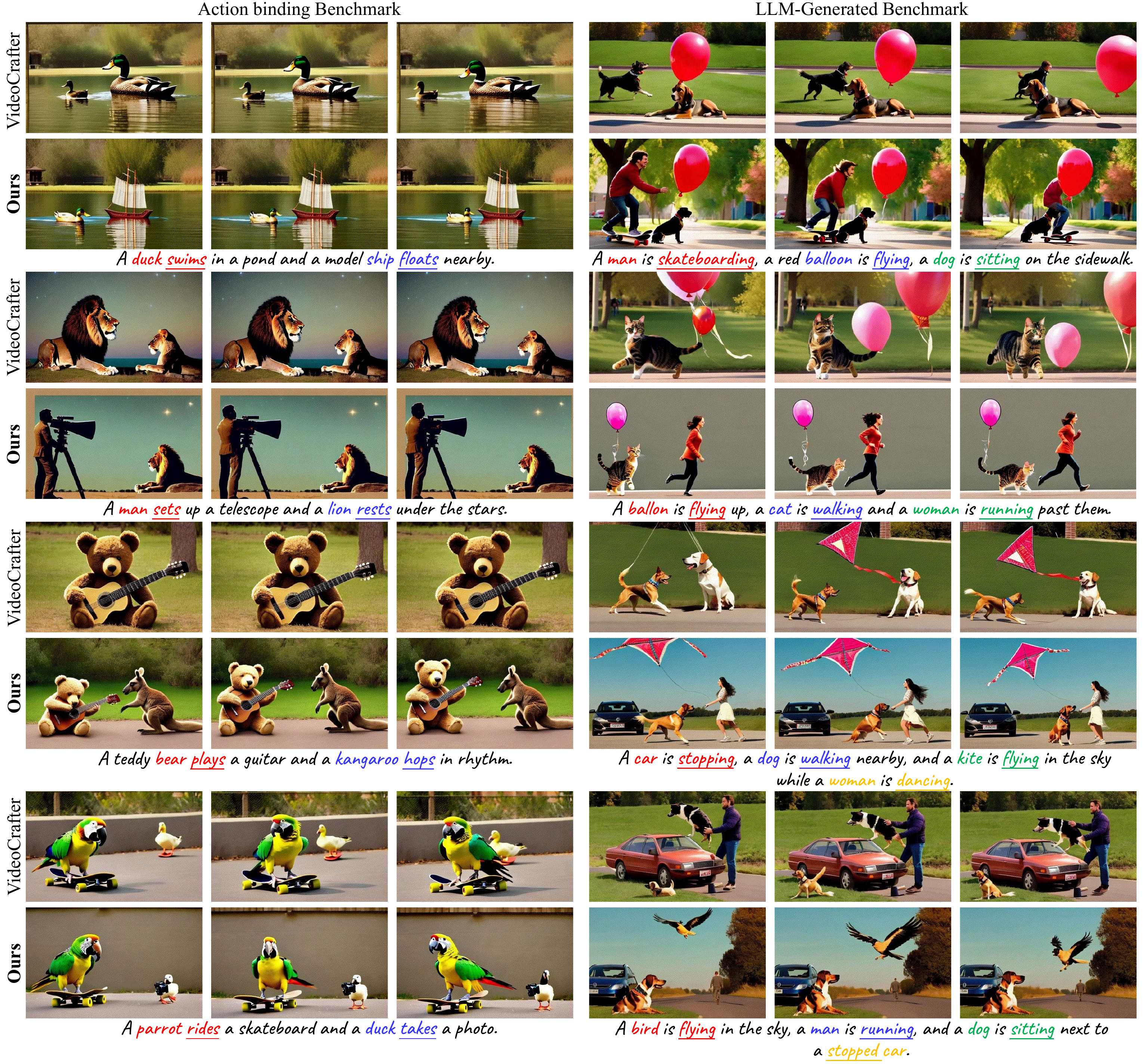}    
    \vspace{-7 mm}
    \caption{\textbf{Qualitative comparisons with VideoCrafter2~\cite{chen2024videocrafter2}.}
    Our proposed \textbf{StarVid} effectively addresses semantic misalignment in VideoCrafter2~\cite{chen2024videocrafter2}.
    }
    \vspace{-6 mm}
    \label{fig:videocrafter_1}
\end{figure*}

\Paragraph{Metrics.}
We evaluate the efficiency of our method in terms of video quality and semantic alignment using the following automatic evaluation metrics:
 1) \textbf{Video quality:} We utilize the Pick-Score~\cite{guo2023animatediff}, which predicts user preferences, and the CLIP Image Similarity (CLIP-I), calculating the cosine similarity between all pairs of video frames using the CLIP~\cite{radford2021learning} image encoder to assess temporal consistency.
 2) \textbf{Semantic alignment:} We first adopt CLIP Text Alignment (CLIP-T), measuring the cosine similarity between video frames and text prompts.
 For our \textbf{compositional setting} with multiple objects, we further leverage advanced metrics from T2V-CompBench~\cite{sun2024t2v}, \ie Numeracy and Action Binding.
 They are designed to evaluate whether the subject's numeracy matches the text description and whether the subject's motion aligns with the text description, respectively.

\begin{figure}[!htp]
    \centering    
    \includegraphics[width=1\linewidth]{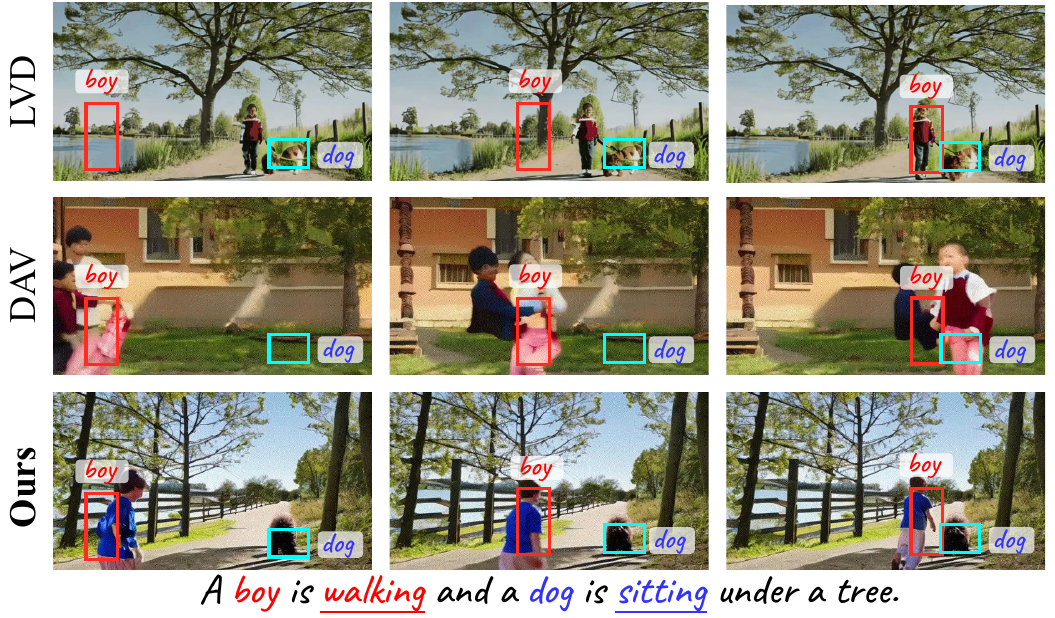}
    \vspace{-7 mm}
    \caption{\textbf{Comparisond of spatial layout adherence between our StarVid and LVD~\cite{lian2023llmgroundedvideo} and DAV~\cite{yang2024direct}.}
    Given the spatial prior, LVD~\cite{lian2023llmgroundedvideo} and DAV~\cite{yang2024direct} cannot be accurately aligned.
     In contrast, our method ensures accurate alignment of the subject within the bounding box.
}
    \label{fig:fig_5}
\end{figure}

\subsection{Comparisons using ZeroScope as Backbone}

To evaluate the performance of our method StarVid, we first conduct experiments using ZeroScope~\cite{wang2023modelscope} as the backbone model.

\Paragraph{Baseline.}
We compare our method with the baseline ZeroScope~\cite{wang2023modelscope} and two other methods that also employ ZeroScope~\cite{wang2023modelscope} as their backbone:
1) \emph{LVD}~\cite{lian2023llmgroundedvideo}, which incorporates the LLM to generate dynamic layouts from text prompts, thereby guiding video generation.
2) \emph{Director-A-Video (DAV)}~\cite{yang2024direct}, which utilizes object motion guidance to control the video generation process.

\Paragraph{Qualitative Results.}
As illustrated in \figref{fig_4}, the baseline ZeroScope~\cite{wang2023modelscope} struggles to generate the correct number of subjects.
While LVD~\cite{lian2023llmgroundedvideo} accurately produces the required number of subjects, it exhibits significant motion-subject misalignment.
For example, in the prompt ``\emph{a \red{man} is \red{\underline{walking}} while a \blue{woman} \blue{\underline{rides}} a horse nearby.}'', the woman is depicted as walking, and the man as riding a horse.
Furthermore, despite using layout to guide subject generation, DAV~\cite{yang2024direct} tends to incorrectly generate objects, such as an umbrella-shaped UFO instead of an airplane.
In contrast, our method accurately generates subjects and aligns their motions with textual descriptions, thanks to our well-designed attention-refocusing techniques.

Additionally, although the videos generated by LVD~\cite{lian2023llmgroundedvideo} and DAV~\cite{yang2024direct} incorporate spatial layouts for guidance, they fail to adhere to the specified spatial positional priors, as illustrated in \figref{fig_5}.
For instance, in the prompt ``\emph{A \red{boy} is \red{\underline{walking}} and a \blue{dog} is \blue{\underline{sitting}} under a tree.}'', the ``\emph{boy}'' produced by LVD~\cite{lian2023llmgroundedvideo} remains stationary, not following the designated path from left to right, and DAV~\cite{yang2024direct} incorrectly places the ``\emph{dog}'' at the specified location.
Compared to LVD~\cite{lian2023llmgroundedvideo} and DAV~\cite{yang2024direct}, our method adheres well to the spatial layout, accurately generating ``\emph{boy}'' that moves from left to right and the ``\emph{dog}'' sitting still.

\begin{table*}[!t]
\centering
\caption{\textbf{Human evaluation results of LLM-Generated benchmark.}
Our method is significantly more preferred by users compared to comparative methods.
}
\vspace{-1 mm}
\resizebox{\linewidth}{!}{%
\begin{tabular}{cccccc}
\hline
\multicolumn{1}{c}{} & \multicolumn{3}{c}{\textbf{Video Quality}} & \multicolumn{2}{c}{\textbf{Semantic Alignment}} \\ \cmidrule(r){2-4} \cmidrule(r){5-6}
\multicolumn{1}{c}{\multirow{-2}{*}{\textbf{Method/Metrics}}} & \textbf{Overall Preference} & \textbf{Video Fluency} & \textbf{Video Quality} & \textbf{Quantity Correctness} & \textbf{Motion Correctness} \\ 
\midrule
  \textbf{Ours} v.s. ZeroScope~\cite{wang2023modelscope} & \textbf{94.0\%} v.s. 6.0\% & \textbf{87.0\%} v.s. 13.0\% & \textbf{87.5\%} v.s. 12.5\% & \textbf{97.0\%} v.s. 3.0\% & \textbf{98.0\%} v.s. 2.0\% \\
  \textbf{Ours} v.s.  LVD~\cite{lian2023llmgroundedvideo} & \textbf{93.5\%} v.s. 6.5\% & \textbf{93.0\%} v.s. 7.0\% & \textbf{89.5\%} v.s. 10.5\% & \textbf{97.0\%} v.s. 3.0\% & \textbf{93.5\%} v.s. 6.5\% \\
  \textbf{Ours} v.s.  DAV~\cite{yang2024direct} & \textbf{98.0\%} v.s. 2.0\% & \textbf{96.5\%} v.s. 3.5\% & \textbf{97.5\%} v.s. 2.5\% & \textbf{98.0\%} v.s. 2.0\% & \textbf{99.0\%} v.s. 1.0\% \\
\midrule
    \textbf{Ours} v.s. VideoCrafter2~\cite{chen2024videocrafter2} & \textbf{87.5\%} v.s. 12.5\% & \textbf{86.7\%} v.s. 13.3\% & \textbf{84.2\%} v.s. 15.8\% & \textbf{94.1\%} v.s. 5.9\% & \textbf{94.1\%} v.s. 5.9\% \\
\bottomrule
\end{tabular}%
}
\vspace{-7 mm}
\label{tab:tab_userstudy}
\end{table*}

\begin{table}[!t]
\centering
\caption{\textbf{Quantitative comparison on proposed constraint.}
The best value is highlighted in \colorbox{pearDark!20}{blue}.
}
\vspace{-1 mm}
\resizebox{\linewidth}{!}{
\begin{tabular}{c|ccccc}
\toprule 
\multicolumn{1}{c}{} & \multicolumn{2}{c}{\textbf{Video Quality}} & \multicolumn{3}{c}{\textbf{Semantic Alignment}}\\ \cmidrule(r){2-3} \cmidrule(r){4-6}
\multicolumn{1}{c}{\multirow{-2}{*}{\textbf{Method/Metrics}}} & \textbf{Pick Score} \textbf{($\uparrow$)} & \textbf{CLIP-I} \textbf{($\uparrow$)} & \textbf{CLIP-T} \textbf{($\uparrow$)} & \textbf{Numeracy} \textbf{($\uparrow$)} & \textbf{Action Binding} \textbf{($\uparrow$)} \\ 
\midrule
w/o \textcolor{sp}{$\mathcal{L}_{sp}$}  & 20.42 & \colorbox{pearDark!20}{0.94} & 25.52 & 0.545 & 0.573  \\
w/o $\mathcal{L}_{bg}$ & 20.64 & \colorbox{pearDark!20}{0.94}  & 27.30 & 0.746 & 0.738 \\
w/o \textcolor{sy}{$\mathcal{L}_{syt}$}  & 20.63 & \colorbox{pearDark!20}{0.94} & 27.74 & 0.847 & 0.776 \\
\textbf{Ours} & \colorbox{pearDark!20}{20.69}  & \colorbox{pearDark!20}{0.94} & \colorbox{pearDark!20}{27.76} &  \colorbox{pearDark!20}{0.871} & \colorbox{pearDark!20}{0.795} \\
\bottomrule 
\end{tabular}
}
\vspace{-2 mm}
\label{tab:ablation}
\end{table}

\begin{figure}[!t]
    \centering
    \includegraphics[width=1\linewidth]{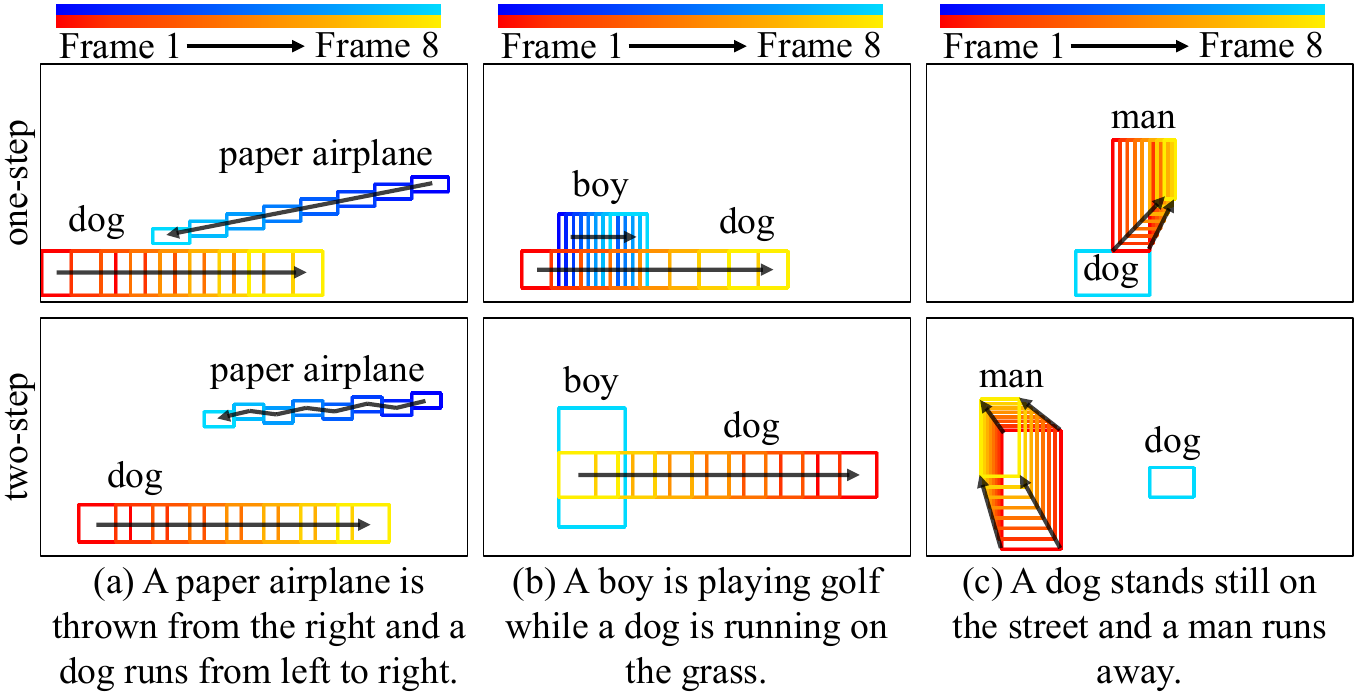} 
    \vspace{-5 mm}
    \caption{\textbf{Motion trajectory planner comparisons.}
    Our motion trajectory planner generates motion trajectories that more closely align with the physical laws of the real world.
    }
    \vspace{-3 mm}
    \label{fig:llm1vs2}
\end{figure}

\Paragraph{Quantitative Results.}
We quantitatively evaluate our proposed
method against baseline models using automatic metrics.
\tabref{tab_1} illustrates that our proposed method outperforms all other baselines in both benchmarks.
Notably, in terms of numeracy correctness and action binding, our method significantly surpasses the comparison methods, demonstrating that the videos generated by our approach effectively enhance semantic alignment in settings involving multiple objects.
Consequently, this improvement also leads to superior performance in the CLIP-T and Pick-Score metrics.

\subsection{Comparisons using VideoCrafter2 as Backbone}
We further employ VideoCrafter2 as the backbone model in this section.
Compared to ZeroScope~\cite{wang2023modelscope}, VideoCrafter2~\cite{chen2024videocrafter2} produces videos with superior visual quality and motion performance.

\Paragraph{Qualitative Results.} \figref{videocrafter_1} presents qualitative results using VideoCrafter2~\cite{chen2024videocrafter2} as the baseline model.
VideoCrafter2~\cite{chen2024videocrafter2} struggles to generate subjects and motions consistent with the text description, while our method enables it to produce correct motions and subjects. 
For example, given the text prompt ``\emph{A \red{duck} \red{\underline{swims}} in a pond and a model \blue{ship} \blue{\underline{floats}} nearby.}'', VideoCrafter2~\cite{chen2024videocrafter2} generates two ducks. 
In contrast, our method successfully generates a video where the ``\emph{duck}'' is swimming and the ``\emph{ship}'' is floating.

\begin{table}[!t]
\centering
\caption{\textbf{Quantitative comparison on Multi-Frame Contrastive Strategy.}
The best value is highlighted in \colorbox{pearDark!20}{blue}.
}
\vspace{-2 mm}
\resizebox{\linewidth}{!}{%
\begin{tabular}{cccccc}
\hline
\multicolumn{1}{c}{} & \multicolumn{2}{c}{\textbf{Video Quality}} & \multicolumn{3}{c}{\textbf{Semantic Alignment}} \\ 
\cmidrule(r){2-3} \cmidrule(r){4-6}
\multicolumn{1}{c}{\multirow{-2}{*}{\textbf{Method/Metrics}}} & \textbf{Pick Score} \textbf{($\uparrow$)} & \textbf{CLIP-I} \textbf{($\uparrow$)} & \textbf{CLIP-T} \textbf{($\uparrow$)} & \textbf{Numeracy} \textbf{($\uparrow$)} & \textbf{Action Binding} \textbf{($\uparrow$)} \\ 
\midrule
w/o frame & 20.59 & \colorbox{pearDark!20}{0.94} & 27.36 & 0.823 & 0.768  \\
\textbf{num=1} & \colorbox{pearDark!20}{20.69} & \colorbox{pearDark!20}{0.94} & \colorbox{pearDark!20}{27.76} & \colorbox{pearDark!20}{0.871} & \colorbox{pearDark!20}{0.795}  \\
num=2 & 20.64 & \colorbox{pearDark!20}{0.94} & 27.47 & 0.803 & 0.763  \\
num=3 & 20.58 & \colorbox{pearDark!20}{0.94} & 27.34 & 0.819 & 0.774  \\
\bottomrule
\end{tabular}%
}
\vspace{-2 mm}
\label{tab:tab_framecount_alignment}
\end{table}
\begin{figure}[!t]
    \centering
    \includegraphics[width=1\linewidth]{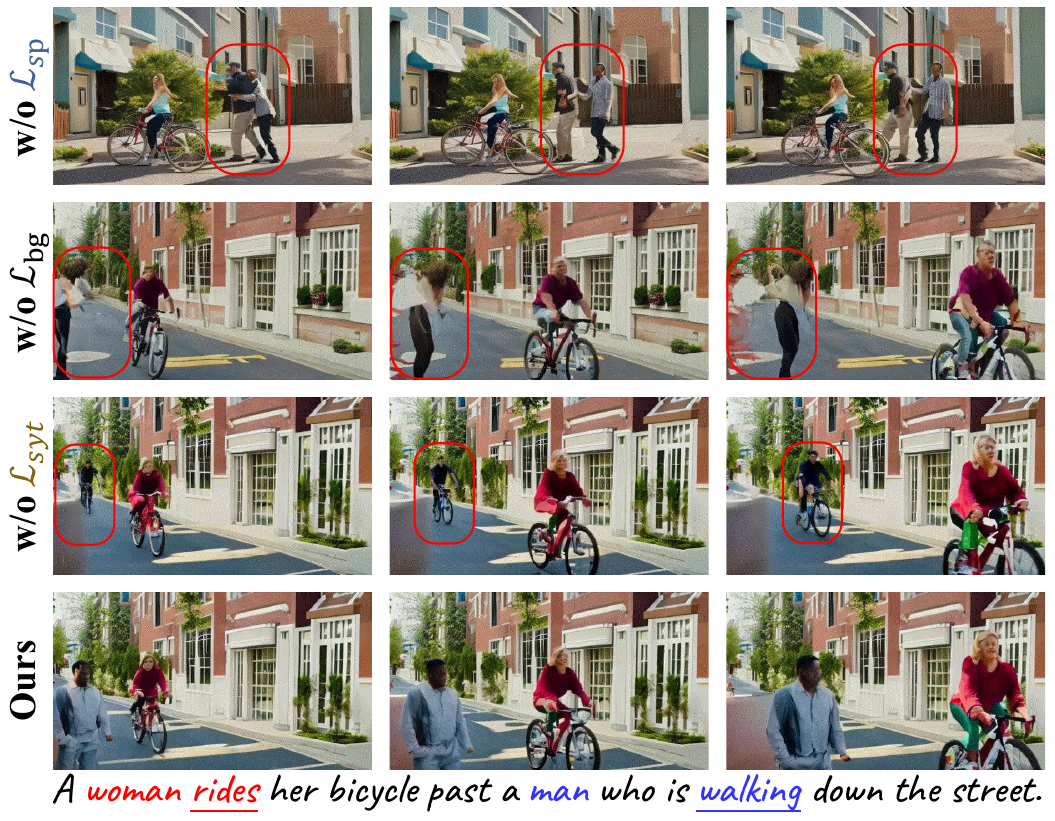}  
    \vspace{-5 mm}
\caption{
\textbf{Ablation study of proposed constraint.}
\textcolor{sp}{$\mathcal{L}_{sp}$} ensures spatial consistency with the motion trajectory,
while the $\mathcal{L}_{bg}$ prevents the foreground from leaking into the background.
Furthermore, the \textcolor{sy}{$\mathcal{L}_{syt}$} is crucial for aligning actions with subjects.} 
\vspace{-2 mm}
    \label{fig:ablation_study}
\end{figure}

\Paragraph{Quantitative Results.} We quantitatively compare the results on VideoCrafter2~\cite{chen2024videocrafter2}.
As illustrated in \tabref{tab_1}, our method significantly outperforms VideoCrafter2~\cite{chen2024videocrafter2} in numeracy correctness and action binding, indicating that the videos generated by our method align more closely with the text prompts.
Additionally, our method also surpasses VideoCrafter2~\cite{chen2024videocrafter2} in CLIP-T and Pick-Score metrics.

\begin{table*}[!t]
\centering
\caption{\textbf{Ablation study of hyper-parameters in spatial-aware and syntax-aware constraints.}
The best value is highlighted in \colorbox{pearDark!20}{blue}.
}
\vspace{-1 mm}
\resizebox{\linewidth}{!}{%
\begin{tabular}{c|cccccc|cccccc}
\hline
\multicolumn{2}{c}{} & \multicolumn{5}{c}{\textbf{Spatial-aware constraint}} & \multicolumn{1}{c}{} & \multicolumn{5}{c}{\textbf{Syntax-aware constraint}} \\ \cmidrule(r){3-7} \cmidrule(r){9-13}
\multicolumn{2}{c}{} & \multicolumn{2}{c}{\textbf{Video Quality}} & \multicolumn{3}{c}{\textbf{Semantic Alignment}} & & \multicolumn{2}{c}{\textbf{Video Quality}} & \multicolumn{3}{c}{\textbf{Semantic Alignment}} \\ \cmidrule(r){3-4} \cmidrule(r){5-7} \cmidrule(r){9-10} \cmidrule(r){11-13} 
\multicolumn{2}{c}{\multirow{-4}{*}{\textbf{Method/Metrics}}} & \multicolumn{1}{c}{\textbf{Pick Score} \textbf{($\uparrow$)}} & \multicolumn{1}{c}{\textbf{CLIP-I} \textbf{($\uparrow$)}} & \multicolumn{1}{c}{\textbf{CLIP-T} \textbf{($\uparrow$)}} & \multicolumn{1}{c}{\textbf{Numeracy} \textbf{($\uparrow$)}} & \multicolumn{1}{c}{\textbf{Action Binding} \textbf{($\uparrow$)}} & \multicolumn{1}{c}{} & \multicolumn{1}{c}{\textbf{Pick Score} \textbf{($\uparrow$)}} & \multicolumn{1}{c}{\textbf{CLIP-I} \textbf{($\uparrow$)}} & \multicolumn{1}{c}{\textbf{CLIP-T} \textbf{($\uparrow$)}} & \multicolumn{1}{c}{\textbf{Numeracy} \textbf{($\uparrow$)}} & \multicolumn{1}{c}{\textbf{Action Binding} \textbf{($\uparrow$)}} \\ 
\midrule
 & timesteps=1 & 20.61 & \colorbox{pearDark!20}{0.94} & 26.42 & 0.639 & 0.650 & timesteps=10 & 20.61 & \colorbox{pearDark!20}{0.94} & 27.68 & 0.835 & 0.755 \\
& timesteps=3 & 20.67 & \colorbox{pearDark!20}{0.94} & 27.35 & 0.772 & 0.736 & \textbf{timesteps=25} & \colorbox{pearDark!20}{20.69} & \colorbox{pearDark!20}{0.94} & \colorbox{pearDark!20}{27.76} & \colorbox{pearDark!20}{0.871} & \colorbox{pearDark!20}{0.795} \\
& \textbf{timesteps=5} & \colorbox{pearDark!20}{20.69} & \colorbox{pearDark!20}{0.94} & \colorbox{pearDark!20}{27.76} & \colorbox{pearDark!20}{0.871} & \colorbox{pearDark!20}{0.795} & timesteps=35 & 20.64 & \colorbox{pearDark!20}{0.94} & 27.52 & 0.837 & 0.765 \\
\multirow{-4}{*}{\rotatebox{90}{Time steps}} & timesteps=7 & 20.53 & \colorbox{pearDark!20}{0.94} & 27.31 & 0.861 & 0.776 & timesteps=50 & 20.52 & \colorbox{pearDark!20}{0.94} & 27.16 & 0.801 & 0.741 \\ 
\midrule
 & iters=5 & 20.66 & \colorbox{pearDark!20}{0.94} & 27.19 & 0.740 & 0.732 & \textbf{iters=1} & \colorbox{pearDark!20}{20.69} & \colorbox{pearDark!20}{0.94} & \colorbox{pearDark!20}{27.76} & \colorbox{pearDark!20}{0.871} & \colorbox{pearDark!20}{0.795} \\
& \textbf{iters=10} & \colorbox{pearDark!20}{20.69} & \colorbox{pearDark!20}{0.94} & \colorbox{pearDark!20}{27.76} & \colorbox{pearDark!20}{0.871} & \colorbox{pearDark!20}{0.795} & iters=2 & 20.48 & \colorbox{pearDark!20}{0.94} & 27.09 & 0.823 & 0.746 \\
& iters=15 & 20.44 & \colorbox{pearDark!20}{0.94} & 27.13 & 0.835 & 0.765 & iters=5 & 20.21 & \colorbox{pearDark!20}{0.94} & 25.73 & 0.785 &  0.701 \\
\multirow{-4}{*}{\rotatebox{90}{Max-iters}} & iters=20 & 20.33 & \colorbox{pearDark!20}{0.94} & 26.91 & 0.832 & 0.736 & iters=10 & 19.97 & \colorbox{pearDark!20}{0.94} & 25.25 & 0.770 & 0.669 \\ 
\midrule
& loss weight=10 & 20.61 & \colorbox{pearDark!20}{0.94} & 26.87 & 0.693 & 0.703  & loss weight=10 & 20.61 &\colorbox{pearDark!20}{0.94} & 27.34 & 0.813 & 0.744 \\
 & loss weight=20 & 20.59 & \colorbox{pearDark!20}{0.94} & 27.13 & 0.741 & 0.736 & \textbf{loss weight=20} & \colorbox{pearDark!20}{20.69} & \colorbox{pearDark!20}{0.94} & \colorbox{pearDark!20}{27.76} & \colorbox{pearDark!20}{0.871} & \colorbox{pearDark!20}{0.795}\\
& \textbf{loss weight=30} & \colorbox{pearDark!20}{20.69} & \colorbox{pearDark!20}{0.94} & \colorbox{pearDark!20}{27.76} & \colorbox{pearDark!20}{0.871} & \colorbox{pearDark!20}{0.795} & loss weight=30 & 20.48 & \colorbox{pearDark!20}{0.94} & 27.19 & 0.863 & 0.739 \\
\multirow{-4}{*}{\rotatebox{90}{\begin{tabular}[c]{@{}c@{}}loss\\ weight\end{tabular}}} & loss weight=40 & 20.51 & \colorbox{pearDark!20}{0.94} & 27.35 & 0.853 &  0.742 & loss weight=40 & 20.31 & \colorbox{pearDark!20}{0.94} & 26.42 & 0.801 & 0.722 \\ 
\bottomrule
\end{tabular}%
}
\vspace{-3 mm}
\label{tab:tab_ablation_a}
\end{table*}

\begin{table}[!t]
\centering
\vspace{-3 mm}
\caption{\textbf{Ablation study of distance functions and formulas for syntax-aware constrains.}
The best value is highlighted in \colorbox{pearDark!20}{blue}.
}
\vspace{-1 mm}
\resizebox{\linewidth}{!}{%
\begin{tabular}{cccccc}
\hline
\multicolumn{1}{c}{} & \multicolumn{2}{c}{\textbf{Video Quality}} & \multicolumn{3}{c}{\textbf{Semantic Alignment}} \\ 
\cmidrule(r){2-3} \cmidrule(r){4-6}
\multicolumn{1}{c}{\multirow{-2}{*}{\textbf{Method/Metrics}}} & \textbf{Pick Score} \textbf{($\uparrow$)} & \textbf{CLIP-I} \textbf{($\uparrow$)} & \textbf{CLIP-T} \textbf{($\uparrow$)} & \textbf{Numeracy} \textbf{($\uparrow$)} & \textbf{Action Binding} \textbf{($\uparrow$)} \\ 
\midrule
w/ cosine distance & 20.64 & \colorbox{pearDark!20}{0.94} & 27.65 & 0.847 & 0.754 \\
\midrule
w/ $\mathcal{L}_{pos}-\mathcal{L}_{neg}$ & 20.62 & \colorbox{pearDark!20}{0.94} & 27.42 & 0.839 & 0.768 \\
w/ InfoNCE & 20.58 & \colorbox{pearDark!20}{0.94} & 27.39 & 0.815 & 0.759 \\
\textbf{Ours} & \colorbox{pearDark!20}{20.69} & \colorbox{pearDark!20}{0.94} & \colorbox{pearDark!20}{27.76} & \colorbox{pearDark!20}{0.871} & \colorbox{pearDark!20}{0.795}  \\
\bottomrule
\end{tabular}%
}
\vspace{-3 mm}
\label{tab:tab_formal_alignment}
\end{table}

\subsection{User Study}
We conduct a user study on the LLM-generated benchmark to gain a better understanding of user preferences, focusing on two key aspects: \emph{video quality} and \emph{semantic alignment}.
In an A/B test, participants are presented with a text prompt and two generated results from different methods. The results are displayed in random order to prevent participants from inferring which video was generated by which algorithm.
To evaluate the quality of videos,  we ask participants three questions:
\begin{compactitem}
\item \textbf{Video Quality}: Which option has the better video quality?
\item \textbf{Video Fluency}: Which option is more coherent and smooth?
\item \textbf{Overall Preference}: Subjectively, which option do you prefer?
\end{compactitem}
For video-text alignment, we need participants to answer the following questions regarding the number of subjects and motion alignment: 
\begin{compactitem}
\item \textbf{Number Correctness}: Which option has the most consistent Number of subjects with the prompt?
\item \textbf{Motion Correctness}: Based on the provided prompt, which option is the appropriate Motion for the subjects?
\end{compactitem}
We randomly select $30$ generated videos from each of the two baseline models. 
We collect responses from $20$ participants between the ages of $20$ and $29$. 
As demonstrated in \tabref{tab_userstudy}, human evaluation results demonstrate that videos generated by our model significantly outperform those from existing frameworks in both visual quality and semantic alignment, particularly regarding the number of subjects and the alignment of motion. 
Specifically, in terms of motion alignment, most participants consider our method superior to the baselines, achieving $98.0\%$ against ZeroScope~\cite{wang2023modelscope} and $94.1\%$ against VideoCrafter2~\cite{chen2024videocrafter2}, with comparison rates of $93.5\%$ and $99.0\%$ relative to LVD~\cite{lian2023llmgroundedvideo} and DAV~\cite{yang2024direct}, respectively.
This result shows that the proposed method better achieves motion correspondence in multi-subject, multi-motion video generation.

\begin{figure}[!t]
    \centering
    \includegraphics[width=1\linewidth]{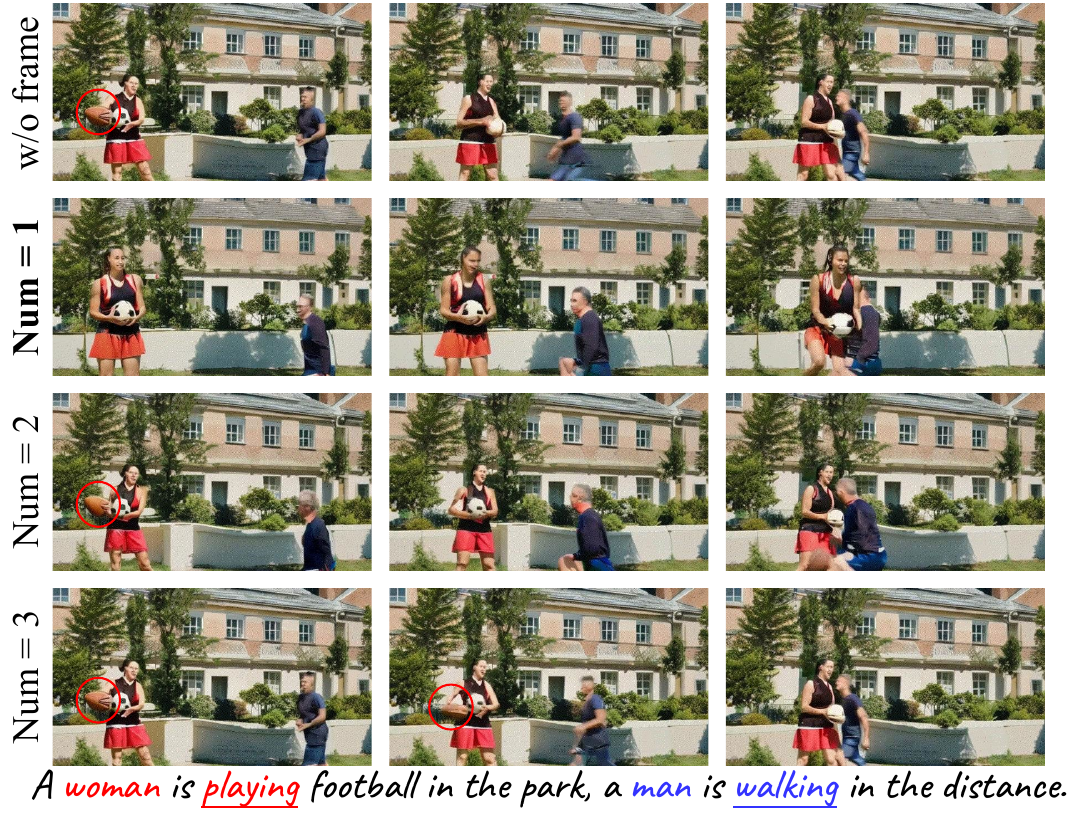}    
    \vspace{-5 mm}
\caption{
\textbf{Ablation study of multi-frame contrastive strategy.}
The multi-frame strategy can prevent the subject from suddenly appearing in a specific frame, where ``w/o frame'' represents ours without the multi-frame contrastive strategy.
} 
\vspace{-3 mm}
    \label{fig:frame_ablation}
\end{figure}

\subsection{Ablation Studies}
We conduct a comprehensive ablation study on the factors outlined below to better understand the proposed approach.

\Paragraph{Motion Trajectory planner.}
\label{sec:llm1vs2}
To evaluate the efficiency of our proposed two-stage motion trajectory planner, we design a baseline using a one-stage LLM planner that generates trajectories in a single step.
%
As illustrated in the \figref{llm1vs2}, our planner generates motion trajectories that adhere more closely to the physical rules of the real world compared to the one-stage planner.
%
For instance, the proposed two-stage planner accounts for the gravity of the subject and incorporates other physical rules, such as aerodynamics (\figref{llm1vs2}(a)), which the single-stage planner cannot manage.
%
In scenarios such as planning human behavior, \eg playing golf, our planner utilizes learned knowledge to accurately plan stationary trajectories (\figref{llm1vs2}(b)) rather than linear movement.
%
Additionally, our planner more accurately plans the perspective geometry of camera motion (\figref{llm1vs2}(c)), reflecting how the subject’s size and the position of the head change as the subject moves away.

\Paragraph{Proposed Attention-Based Guidance.}
We conduct both quantitative and qualitative analyses on the impact of different constraints in StarVid, as shown in \figref{ablation_study} and \tabref{ablation}.
The absence of \textcolor{sp}{$\mathcal{L}_{sp}$} compromises the alignment of subjects into distinct regions, leading to incorrect numeracy and resulting in the lowest values for numeracy and action binding.
Additionally, the exclusion of $\mathcal{L}_{bg}$ in \textcolor{sp}{$\mathcal{L}_{sp}$} leads to foreground leakage into the background, causing the model to mistakenly generate the foreground man as blended with the background, as illustrated in the second row.
The third row further demonstrates that incorporating \textcolor{sy}{$\mathcal{L}_{syt}$} prevents motion leakage, ensuring the man is depicted running rather than riding.

%
%
\begin{figure}[!t]
    \centering
    \includegraphics[width=1\linewidth]{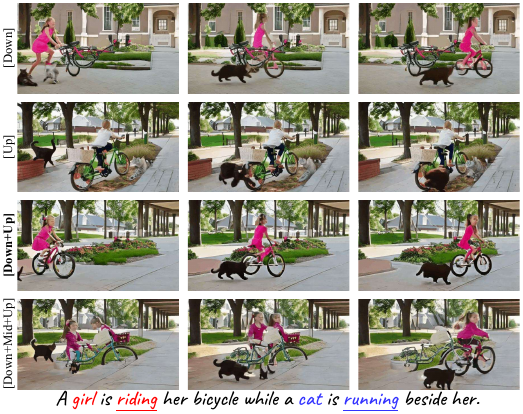} 
    \vspace{-5 mm}
    \caption{\textbf{Ablation study of CAMap's layers.}
    The combination of upsampling and downsampling achieves satisfactory results.
    }
    \vspace{-2 mm}
    \label{fig:UNet_layer}
\end{figure}

\Paragraph{Multi-Frame Contrastive Strategy.}
\figref{frame_ablation} demonstrates that multi-frame contrastive strategy enhances motion consistency, particularly by preventing the subject from abruptly appearing in a specific frame. 
However, increasing the number of adjacent frames does not mitigate this phenomenon; instead, it exacerbates the issue, as confirmed by the results in \tabref{tab_framecount_alignment}.

\Paragraph{The Impact of Hyper-parameters in Spatial-Aware Constraint.}
To investigate the effects of hyperparameters in spatial-aware constraint, we set the range of time steps to \{$1$, $3$, $5$, $7$\}, the range for the maximum number of iterations per time step to \{$5$, $10$, $15$, $20$\}, and the range of loss weights to \{$10$, $20$, $30$, $40$\}. 
As shown in \tabref{tab_ablation_a}, an overly small hyperparameter value leads to a significant reduction in numeracy, resulting in an inconsistency between the number of generated subjects and the text prompt. 
Conversely, an overly large value results in a substantial decrease in Pick-Score, which compromises video quality. 
Based on the automated metrics in \tabref{tab_ablation_a} and comprehensive evaluation of performance and efficiency, we select the timesteps,  the maximum number of iterations, and the loss weights of the spatial-aware constraint as $5$, $10$ and $30$, respectively.

\Paragraph{The Impact of Hyper-parameters in Syntax-Aware Constraint.}
We quantitatively analyze the influence of syntax-aware hyperparameters on the results, as demonstrated in \tabref{tab_ablation_a}.
For syntax-aware constraint, setting excessively small values for the time step and loss weight cause a decrease in the action binding index, indicating the presence of motion leakage. 
In contrast, setting excessively high values can degrade video quality, as evidenced by the decrease in the Pick-Score index and the action binding index. 
Regarding the maximum number of iterations per time step for syntax-aware constraints, the highest automatic index is obtained in one iteration, and the various indexes gradually decrease as the number of iterations increases.
Therefore, we determine the time steps as $25$, the maximum number of iterations as $1$, and the loss weight as $20$, respectively.

\Paragraph{Selection of Distance Functions.}
As shown in \tabref{tab_formal_alignment}, we employ the cosine distance and the KL divergence to measure the distance between CA Maps.
In contrast, KL divergence effectively establishes the connection between subject and motion, significantly outperforming the cosine distance in both numeracy accuracy and action binding. 
Consequently, we adopt the KL divergence to measuree the distance between CA Maps.

\Paragraph{Performance of Different Formulation.}
We conduct an ablation experiment to evaluate the effectiveness of the loss function \eqnref{L_camapc} within the syntax-aware constraint.
The quantitative results in \tabref{tab_formal_alignment} demonstrate that altering the form of \eqnref{L_camapc} to $\mathcal{L}_{pos}-\mathcal{L}_{neg}$ or \textbf{InfoNCE} results in decreased performance in terms of numeracy correctness and action binding. 
By contrast, our design of \eqnref{L_camapc} demonstrates superior performance, achieving the highest score and enabling the generation of high-quality video results that align well semantically with the textual prompts.

\Paragraph{Get CA Maps from which Layer. }
Previous studies~\cite{hertz2022prompt, chen2024training, xie2023boxdiff} have indicated that the $16 \times 16$ and $8 \times 8$ resolution cross-attention layers in the denoising UNet influence the subject's shape and video layout. 
In this section, we perform an ablation experiment to identify which cross-attention layers at these resolutions contribute to enhanced semantic alignment. 
The results are shown in \figref{UNet_layer}.
As depicted in the first two rows of \figref{UNet_layer}, employing only upsampling or downsampling layers leads to noisy background and foreground information, motion loss, and an increase in subject matter.
Additionally, the last row illustrates the outcome of combining upsampling, downsampling, and intermediate layers, which results in the appearance of multiple ``\emph{girls}'' and ``\emph{bicycles}''.
Combining up-sampling and down-sampling layers yields satisfactory results; thus, we obtain the CA Maps from the lowest resolution layer that includes both up-sampling and down-sampling.

\section{Conclusions and Future Works}
In this work, we introduce \textbf{StarVid}, a plug-and-play method designed to enhance the semantic alignment of generated videos when text prompts involve multiple subjects with distinct motions.
Using the spatial reasoning capabilities of LLMs, we design a two-stage motion trajectory planner that generates spatial layout guidance, enabling two proposed attention-refocusing constraints to precisely position subjects and connect their motions.
Extensive experiments demonstrate that our method significantly outperforms baseline approaches, achieving improved semantic consistency, particularly in terms of numerical accuracy and motion binding.

However, a primary limitation lies in the increased inference time introduced by the latent optimization process.
Additionally, the quality of video results depends on the performance of the baseline model.
In future work, we aim to explore acceleration techniques and assess the applicability of our method to a more powerful backbone.

\clearpage
\bibliographystyle{IEEEtran}
\bibliography{main}

\clearpage
\newpage
\appendix

\setcounter{page}{1}
%

\subsection{Summary}

In this supplementary material, we provide detailed implementation details, ablation study results, and additional findings as follows:

\begin{compactitem}
\item In \secref{supp_mtp}, we present the implementation details of the proposed motion trajectory planner.
\item In \secref{supp_data}, we provide a detailed introduction to the benchmark datasets, baselines, and quantitative metrics used in our experiments.
\item We provide additional qualitative results from ablation experiments in Section \secref{motion_ablation}.
\item In \secref{supp_cost}, we analyze the computational overhead of the proposed method.
\item In \secref{supp_addr}, we provide further qualitative results comparing the two backbones on two benchmarks.
\end{compactitem}

\subsection{Details of Motion Trajectory Planner}
\label{sec:supp_mtp}
Our two-stage motion planning approach leverages the logical reasoning and spatial planning capabilities of large language models (LLMs), \ie GPT-4o~\cite{gpt4o}, to incrementally plan motion trajectories from textual prompts that are consistent with real-world physical laws.

\Paragraph{Subject and Motion Reasoning.} 
To enable subject and motion reasoning, as outlined in \tabref{llm_1_task}, we instruct the LLM to identify the subject, the number of subjects, and the motion described in the text prompt. 
The LLM is then instructed to generate the motion plan and provide an explanation for its reasoning.
To ensure that the motion plans generated by the LLM comply with real-world physical laws and that the parsing results are consistent with the text prompts, we provide the LLM with three context examples, as demonstrated in \tabref{llm_1_IoC}.

\Paragraph{Motion Trajectory Prediction.} 
To enable the LLM to plan a motion trajectory that aligns with real-world physical laws, we provide the LLM with task objectives and associated rules, as shown in \tabref{llm_2_task}.
Furthermore, to ensure that the motion trajectory planned by the LLM adheres to real-world physical laws, we provide three context examples, as demonstrated in \tabref{llm_2_IoC}.
These examples include linear motion, nonlinear motion, camera motion, and \etc.

\subsection{Details of Comparisons with Baselines}
\label{sec:supp_data}

\subsubsection{Benchmark Dataset}
\vspace{-1 mm}
To generate text prompts featuring various motions and subjects, we collect common subjects (\eg ``man'', ``woman'', ``dog'', ``cat'', ``car'') and motions (\eg ``running'', ``standing'', ``flying'', ``skateboarding'') for video generation.
Since video generation requires specific prompts, we exclude overly abstract subjects and motions, such as ``human'', ``meditation'' and ``thinking''. 
The final compilation of subjects and motions is as follows:
\begin{compactitem}
\item \textbf{Human Subjects:} Man, Woman, Boy, Girl, Robot
\item \textbf{Animals:} Dog, Cat, Tiger, Bear, Lion, Elephant, Bird, Horse, Cow, Sheep, Dolphin, Fish
\item \textbf{Objects:} Football, Basketball, Car, Motorcycle, Tank, Airplane, Kite, Balloon, Boat
\item \textbf{Linear Actions:} Running, Walking, Skateboarding, Flying, Riding bicycle, Swimming, Driving, Riding horse, Sailing
\item \textbf{Non-Linear Actions:} Jumping, Bouncing, Playing golf, Weightlifting, Playing guitar, Playing football, Dancing, Diving, Sitting (Standing or Stopping)
\end{compactitem}

We select appropriate subject-motion pairs from the collected subjects and motions, excluding unreasonable combinations, such as ``elephant flying'' and ``fish running''. 
We use employ LLM to generate the \textbf{LLM-Generated Benchmark} based on the subject-motion pairs.

To mimic human language patterns, we employ LLM (\ie GPT-4o~\cite{gpt4o}) to generate text prompts based on subject-motion pairs.
We utilize the language reasoning capabilities of LLM to generate text prompts that align with human language patterns from selected subject-motion pairs.    
To enable LLM to generate suitable text prompts based on input subject-motion pairs, we design a unique prompt, as illustrated in \tabref{llm_generated_task}.
Additionally, to accurately guide LLM in understanding our requirements, we provide five contextual examples, as shown in \tabref{llm_generated_IoC}.
For each subject-motion pair, we randomly select one of the four generated prompts as the final text prompt.
Finally, we create 200 text prompts for the LLM-Generated Benchmark.
Unlike the Action Binding Benchmark, \emph{the LLM-Generated Benchmark features a broader range of subjects and their corresponding motions.} 
Examples of text prompts from the LLM-Generated Benchmark include:
\begin{compactitem}
\item A woman is weightlifting, while a man rides a horse in the background.
\item A robot is standing still, while a dog is running in circles. 
\item A boy is walking, and a dog is sitting under a tree.
\item A car stops by the road as an airplane flies across the sky.
\item On a dusty trail, a jeep is driving while a motorcycle halts to the side.
\item The jeep is driving down the road, while a man stands still on the sidewalk.
\item An airplane is flying in the sky, and a woman is running in the park.
\item A dog is sitting quietly, a man is walking ahead, and an airplane flies in the sky.
\item A woman is walking on the path, a dog runs beside her, and a bird is flying overhead
\item A kite is flying in the bright sky while a girl runs below, a boy stands still, and a cat walks along the sidewalk.
\end{compactitem}

\begin{table*}[!t]
\setlength\tabcolsep{0pt}
\caption{Our prompt for subject and motion reasoning task.}
\centering
\begin{tabular*}{\linewidth}{@{\extracolsep{\fill}} l }\toprule
\begin{lstlisting}[style=myverbatim]
You are an expert in extracting relevant information from text prompt. Given a text prompt for generating a video, you will analyze the subjects, number of subjects, and motion of the subjects contained in the text prompt. You will plan how the subjects will move in the video and provide a concise reasoning statement for such planning, no longer than a few sentences. Your response should be in the form of '[{'id': unique object identifier incrementing from 0, 'subject': subject name, 'number': the number of subjects, 'motion': the motion of the subject,}, {'motion planner': The movement of the subject in the video, 'reasoning': The reason for planning this way}]'.  Motion planning should avoid subject overlap. Refer to the examples below for the desired format. Never use markdown or other formats not in the examples. Do not start each frame with '-'. Do not include any comments in your response. 
 
[in-context examples]
prompt: {User text prompt for video generation} 
content: [{'id': unique object identifier incrementing from 0, 'subject': subject name, 'number': the number of subject, 'motion': the motion of subject,}],[{'motion planner': The movement of the subject in the video, 'reasoning': The reason for planning this way}]'
\end{lstlisting} \\\bottomrule
\end{tabular*}
\label{tab:llm_1_task}
\end{table*}

\begin{table*}[!t]
\setlength\tabcolsep{0pt}
\caption{Our in-context examples for subject and motion reasoning task.}
\centering
\begin{tabular*}{\linewidth}{@{\extracolsep{\fill}} l }\toprule
\begin{lstlisting}[style=myverbatim]
prompt: {A kite flies in the sky, and a ball is bouncing to the ground.}
context:[{'id':0, 'subject': kite, 'number': 1, 'motion': flies,}, {'id': 1, 'subject': ball, 'number': 1, 'motion': bouncing,}],{'motion planner': The kite flies from the top of the picture to the upper left corner. The ball bounces back and forth on the right side of the picture. 'reasoning': The kite moves from the top of the frame to the upper left, so its y coordinate remains constant while its x coordinate decreases. The ball bounces on the right, so its x coordinate should remain constant, its y coordinate should increase, and its speed should be faster in later frames until it hits the ground, at which point it bounces back due to its elasticity.}]

prompt: {A dog is walking towards the camera, a cat is sitting, zoom out.}
content: [{'id': 0, 'subject': dog,'number': 1,'motion': walking,}, {'id': 1, 'subject': cat,'number': 1,'motion': sitting,}],{'motion planner': The dog is on the left side of the frame, chasing the camera. The cat is on the right side of the frame, staying still. 'reasoning':Due to perspective geometry, the dog remains the same size as it moves towards the camera. The cat is sitting, staying still, and getting smaller as you zoom out.}]

prompt: {On the grass, a man is playing-golf and a boy rides a bicycle}
content: [{'id': 0, 'subject': man,'number': 1,'motion': playing-golf,}, {'id': 1, 'subject': man,'number': 1,'motion': rides,}],{'motion planner': The man is playing golf on the left side of the screen, and the boy is riding a bicycle from the middle of the screen to the right side of the screen. 'reasoning':The man is playing golf, and his coordinates remain unchanged. The boy moves from the screen to the right, and his x coordinate gradually increases, while his y coordinate remains unchanged.}]
\end{lstlisting} \\\bottomrule
\end{tabular*}
\label{tab:llm_1_IoC}
\end{table*}

\begin{table*}[!t]
\setlength\tabcolsep{0pt}
\caption{Our prompt for motion trajectory prediction task.}
\centering
\begin{tabular*}{\linewidth}{@{\extracolsep{\fill}} l }\toprule
\begin{lstlisting}[style=myverbatim]
You are an intelligent bounding box generator for videos. You don't need to generate the videos themselves but need to generate the bounding boxes. I will provide you with a video with 8 frames, 4 frames per second, with textual prompts containing the subject, subject motion, subject motion plan and reasoning.  Your task is to generate a list of ground truth bounding boxes for each object. The size of the video frame is 320*576. The top-left corner has coordinates [0, 0]. The bottom-right corner has coordinates [576, 320]. Each frame should be represented as '[{'id': unique object identifier incrementing from 0, 'name': object name, 'box': [box top-left x-coordinate, box top-left y-coordinate, box width, box height]}, ...]'. 
You should follow these rules when generating a list of ground truth bounding boxes:
1. box top-left x-coordinate add box width is less than 576, box top-left y-coordinate add box height is less than 320.
2. Each box should not include more than one object. 
3. Each object's box should not overlap in the same frame.
4. Your generated frames must encapsulate the whole scenario depicted by the caption. 
5. Assume objects move and interact based on real-world physics, considering aspects such as gravity and elasticity. 
6. Assume the camera follows perspective geometry. 
7. Boxes for an object should have the same id across the frames, even if the object may disappear and reappear. 
Refer to the examples below for the desired format. Never use markdown or other formats not in the examples. Do not start each frame with '-'. Do not include any comments in your response. 
\end{lstlisting} \\\bottomrule
\end{tabular*}
\label{tab:llm_2_task}
\end{table*}

\begin{table*}[!t]
\setlength\tabcolsep{0pt}
\caption{Our in-context examples for motion tarjectory prediction task.}
\centering
\begin{tabular*}{\linewidth}{@{\extracolsep{\fill}} l }\toprule
\begin{lstlisting}[style=myverbatim]
{prompt: {A kite flies in the sky, and a ball is bouncing to the ground.} 
content: {Frame 1: [{'id': 0, 'name': 'kite', 'box': [280, 20, 50, 50]}, {'id': 1, 'name': 'ball', 'box': [380, 160, 40, 40]}]
Frame 2: [{'id': 0, 'name': 'kite', 'box': [240, 20, 50, 50]}, {'id': 1, 'name': 'ball', 'box': [380, 180, 40, 40]}]
Frame 3: [{'id': 0, 'name': 'kite', 'box': [200, 20, 50, 50]}, {'id': 1, 'name': 'ball', 'box': [380, 220, 40, 40]}]
Frame 4: [{'id': 0, 'name': 'kite', 'box': [160, 20, 50, 50]}, {'id': 1, 'name': 'ball', 'box': [380, 260, 40, 40]}]
Frame 5: [{'id': 0, 'name': 'kite', 'box': [120, 20, 50, 50]}, {'id': 1, 'name': 'ball', 'box': [380, 240, 40, 40]}]
Frame 6: [{'id': 0, 'name': 'kite', 'box': [80, 20, 50, 50]}, {'id': 1, 'name': 'ball', 'box': [380, 210, 40, 40]}]
Frame 7: [{'id': 0, 'name': 'kite', 'box': [40, 20, 50, 50]}, {'id': 1, 'name': 'ball', 'box': [380, 240, 40, 40]}]
Frame 8: [{'id': 0, 'name': 'kite', 'box': [0, 20, 50, 50]}, {'id': 1, 'name': 'ball', 'box': [380, 260, 40, 40]}]}

{prompt: {A dog is walking towards the camera, a cat is sitting, zoom out.}
content: {Frame 1: [{'id': 0, 'name': 'dog', 'box': [150, 200, 40, 60]}, {'id': 1, 'name': 'cat', 'box': [380, 220, 50, 40]}]
Frame 2: [{'id': 0, 'name': 'dog', 'box': [150, 200, 40, 60]}, {'id': 1, 'name': 'cat', 'box': [377, 218, 45, 36]}]
Frame 3: [{'id': 0, 'name': 'dog', 'box': [150, 200, 40, 60]}, {'id': 1, 'name': 'cat', 'box': [374, 216, 40, 32]}]
Frame 4: [{'id': 0, 'name': 'dog', 'box': [150, 200, 40, 60]}, {'id': 1, 'name': 'cat', 'box': [371, 214, 35, 28]}]
Frame 5: [{'id': 0, 'name': 'dog', 'box': [150, 200, 40, 60]}, {'id': 1, 'name': 'cat', 'box': [368, 212, 30, 24]}]
Frame 6: [{'id': 0, 'name': 'dog', 'box': [150, 200, 40, 60]}, {'id': 1, 'name': 'cat', 'box': [365, 210, 25, 20]}]
Frame 7: [{'id': 0, 'name': 'dog', 'box': [150, 200, 40, 60]}, {'id': 1, 'name': 'cat', 'box': [362, 208, 20, 16]}]
Frame 8: [{'id': 0, 'name': 'dog', 'box': [150, 200, 40, 60]}, {'id': 1, 'name': 'cat', 'box': [359, 206, 15, 12]}]}

prompt: {On the grass, a man is playing-golf and a boy rides a bicycle}
content:{Frame 1: [{'id': 0, 'name': 'man', 'box': [100, 100, 110, 180]}, {'id': 1, 'name': 'boy', 'box': [280, 180, 65, 100]}]
Frame 2: [{'id': 0, 'name': 'man', 'box': [100, 100, 110, 180]}, {'id': 1, 'name': 'boy', 'box': [310, 180, 65, 100]}]
Frame 3: [{'id': 0, 'name': 'man', 'box': [100, 100, 110, 180]}, {'id': 1, 'name': 'boy', 'box': [340, 180, 65, 100]}]
Frame 4: [{'id': 0, 'name': 'man', 'box': [100, 100, 110, 180]}, {'id': 1, 'name': 'boy', 'box': [370, 180, 65, 100]}]
Frame 5: [{'id': 0, 'name': 'man', 'box': [100, 100, 110, 180]}, {'id': 1, 'name': 'boy', 'box': [400, 180, 65, 100]}]
Frame 6: [{'id': 0, 'name': 'man', 'box': [100, 100, 110, 180]}, {'id': 1, 'name': 'boy', 'box': [430, 180, 65, 100]}]
Frame 7: [{'id': 0, 'name': 'man', 'box': [100, 100, 110, 180]}, {'id': 1, 'name': 'boy', 'box': [460, 180, 65, 100]}]
Frame 8: [{'id': 0, 'name': 'man', 'box': [100,100, 110, 180]}, {'id': 1, 'name': 'boy', 'box': [490, 180, 65, 100]}]}
\end{lstlisting} \\\bottomrule
\end{tabular*}
\label{tab:llm_2_IoC}
\end{table*}

\begin{table*}[!t]
\setlength\tabcolsep{0pt}
\caption{Our prompt for automatically generating text prompts using LLM.}
\centering
\begin{tabular*}{\linewidth}{@{\extracolsep{\fill}} l }\toprule
\begin{lstlisting}[style=myverbatim]
You are a large language model, trained on a massive dataset of text. You can generate texts from given examples. You are asked to generate similar examples to the provided ones and follow these rules:
1. You generate the correct description based on the provided subject and subject motion. The provided subject and subject motion is in the form of [{'object', 'motion'},{'object', 'motion'},...]
2. Your generation will be served as prompts for Text-to-Video models. So your prompt should be as visual as possible.
3. Do NOT generate scary prompts.
4. Do NOT repeat any existing examples.
5. Your generated examples should be as creative as possible.
6. Your generated examples should not have repetition.
7. Your generated examples should be as diverse as possible.
8. Do NOT include extra texts such as greetings.
9. Generate 4 descriptions.
10. The descriptions you generate should have a diverse word count, with both long and short lengths.
11. Keep the video description as brief as possible.
12. The length of each sentence is limited to 40 characters.
Please open your mind based on the theme [{'object', 'motion'},{'object', 'motion'},...].
\end{lstlisting} \\\bottomrule
\end{tabular*}
\label{tab:llm_generated_task}
\end{table*}

\begin{table*}[!t]
\setlength\tabcolsep{0pt}
\caption{Our in-context examples for automatically generating text prompts using LLM.}
\centering
\begin{tabular*}{\linewidth}{@{\extracolsep{\fill}} l }\toprule
\begin{lstlisting}[style=myverbatim]
 Here are five example descriptions:
[{'man', 'skateboarding'}, {'dog', 'running'}]:
1. A man is skateboarding and a dog is running.
2. a dog is running and a man is skateboarding on the street.
3. A man is skateboarding on the lawn while his white dog is running.
4. A dog is running on the grass, and not far away, a man is skateboarding.

[{'cat', 'sitting'}, {'bird', 'flying'}]:
1. A cat is sitting still in the corner, a bird is flying in the sky above its head.
2. A cat is sitting and a bird is flying.
3. A bird is flying over the grass and a cat sitting there.
4. A cat sits on the lawn, a bird is flying.

[{'airplane', 'flying'}, {'tiger', 'jumping'}]:
1. An airplane is flying to the right, and a tiger is jumping below.
2. A tiger is jumping on the grassland, and an airplane is flying to the left above its head.
3. A tiger is jumping and an airplane is flying far away.
4. A tiger is jumping on the lawn, an airplane is flying.

[{'man','walking'},{'cat','sitting'},{'bird','flying'}]:
1. A man is walking, a bird is flying in the sky above his head, and a cat is sitting still in the corner.
2. A man is walking and a cat is sitting and a bird is flying.
3. A bird is flying over the grass and a man is walking towards a cat sitting there.
4. A cat sits on the lawn, a bird is flying, and a man is walking into the distance

[{'man','walking'},{'cat','sitting'},{'bird','flying'},{'car','stopping'}]:
1. A man is walking towards a stopped car on the left, a bird is flying in the sky above his head, and a cat is sitting quietly in the corner.
2. A man is walking, a car is stopping, a cat is sitting and a bird is flying.
3. A bird is flying over the grass, a man is walking towards a cat sitting there, and a car is stopped in the distance.
4. A car stopped at the edge of the lawn, a cat sat on the lawn, a bird was flying, and a man was walking in the distance.

Please imitate the above examples to generate diverse text descriptions, and do not repeat the above examples. Each description is intended to vividly convey a smooth-motion video with multiple subjects.

The format of your answer should be:{ " descriptions ":[...] }Ensure that the response can be parsed by json.loads in Python, for example: no trailing commas, no single quotes, and so on.
\end{lstlisting} \\\bottomrule
\end{tabular*}
\label{tab:llm_generated_IoC}
\end{table*}

\subsubsection{Implementation Details of Baselines}
We use the official codes released by authors for Zeroscope~\cite{wang2023modelscope}, LVD~\cite{lian2023llmgroundedvideo}, and DAV~\cite{yang2024direct}.
For DAV~\cite{yang2024direct}, we only use the subject motion control component.
To facilitate fair comparisons, LVD~\cite{lian2023llmgroundedvideo}, DAV~\cite{yang2024direct}, and our method all use the same motion trajectory generated by the motion trajectory planner.

\subsubsection{Evaluation Details}

For each text prompt, we generate two videos, resulting in a total of $400$ per benchmark for evaluation.
We use the CLIP Vit-L/14\footnote{\url{https://huggingface.co/openai/clip-vit-large-patch14}} model to calculate CLIP Image Similarity and CLIP Text Alignment,  assessing the quality of the generated videos in terms of both temporal consistency and alignment with the textual prompts. 
Additionally, we use the PickScore\_v1 model \footnote{\url{https://huggingface.co/yuvalkirstain/PickScore_v1}} to calculate the Pick Score, which evaluates the overall quality and relevance of the generated videos.
%
For specific tasks like evaluating generative numeracy and action binding, we employ the official codes from T2V-ComBench~\cite{sun2024t2v}.
 Following the T2V-ComBench setup, we use the Ground-DINO model (version groundingdino\_swint\_ogc \footnote{\url{https://huggingface.co/ShilongLiu/GroundingDINO/tree/main}}) for assessing generative numeracy, and the LLava model (version llava-v1.6-34b \footnote{\url{https://huggingface.co/liuhaotian/llava-v1.6-34b}}) for evaluating action binding.

\subsection{Additional ablation study results}
\label{sec:motion_ablation}

\subsubsection{The Impact of Hyper-parameters in Spatial-Aware Constraint.}
\begin{figure*}[!t]
    \centering
    \includegraphics[width=1\linewidth]{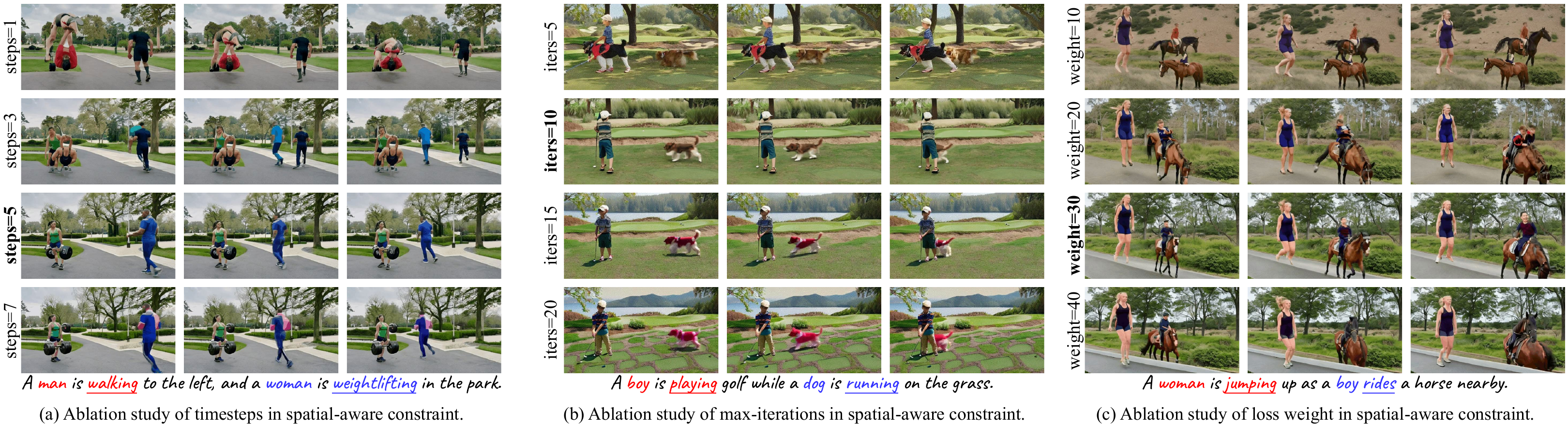}   
    \vspace{-5 mm}
    \caption{\textbf{The impact of hyper-parameters in spatial-aware constraint.}
    Appropriate hyperparameters facilitate the generation of results that are coherent with text semantics.
    }
    \label{fig:fig_ablation_a}
\end{figure*}

\figref{fig_ablation_a}demonstrates the qualitative analysis of how spatial-aware hyperparameters impact performance.
As shown in \figref{fig_ablation_a}, excessively small hyperparameter values cause inconsistency between the number of generated subjects and text prompts (the first two rows of \figref{fig_ablation_a} (a), the first row of \figref{fig_ablation_a} (b) and the first two rows of \figref{fig_ablation_a} (c)), whereas excessive values induce chromatic distortions in subjects (the last two rows of \figref{fig_ablation_a} (b)) or their unintended coalescence (the last row of \figref{fig_ablation_a} (b)). 

\begin{figure*}[!t]
    \centering
    \includegraphics[width=1\linewidth]{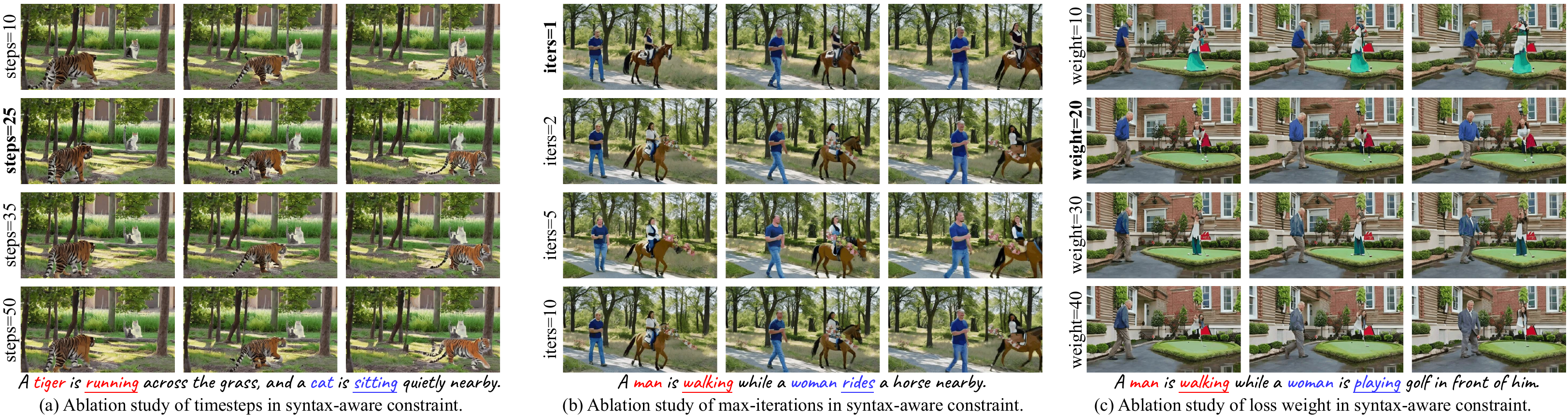} 
    \vspace{-5 mm}
    \caption{\textbf{The impact of hyper-parameters in syntax-aware constraint.}
    Appropriate hyperparameters are crucial for avoiding motion leakage.
    }
    \label{fig:fig_ablation_b}
\vspace{-2 mm}
\end{figure*}

\begin{figure}[!t]
    \centering
    \includegraphics[width=1\linewidth]{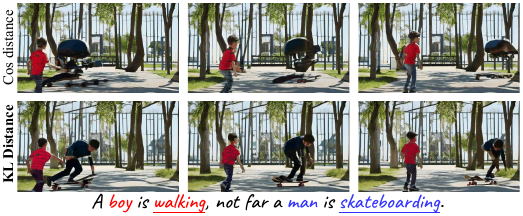}   
    \vspace{-5 mm}
    \caption{\textbf{Ablation study of distance function.}}
    \label{fig:cosvskl}
\end{figure}

\subsubsection{The Impact of Hyper-parameters in Syntax-Aware Constraint.}

We conduct a qualitative analysis of how syntax-aware hyperparameters influence the results, as demonstrated in \figref{fig_ablation_b}.
For syntax-aware constraint, setting excessively small values for the time step and loss weight may cause motion leakage (the first row of \figref{fig_ablation_b} (a) and the first row of \figref{fig_ablation_b} (c)). 
In contrast, setting excessively high values can degrade video quality, leading to deformation (the last two rows of \figref{fig_ablation_b} (a)) and blurring (the last two rows of \figref{fig_ablation_b} (c)) of the subject. 
The maximum number of iterations per time step for the syntax-aware constraint, as shown in \figref{fig_ablation_b} (b), typically requires only a single iteration to achieve accurate results.
However, an excessive number of iterations significantly impairs video quality, resulting in severe artifacts in the last three rows of the figure.

\subsubsection{Selection of Distance Functions.}

As shown in \figref{cosvskl}, using cosine distance results in an inaccurate representation of the ``\emph{man}''. 

In contrast, KL divergence effectively establishes the connection between subject and motion, producing video results that are semantically consistent with text prompts.

\subsubsection{Performance of Different Formulation.}

\begin{figure}[!t]
    \centering
    \includegraphics[width=1\linewidth]{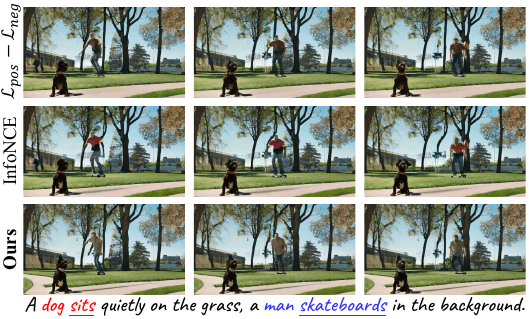}  
    \vspace{-5 mm}
    \caption{\textbf{Comparison of formulas for syntax-aware constrains.}}
    \label{fig:our_vs_infonce_vs_add}
\end{figure}

The qualitative results are illustrated in \figref{our_vs_infonce_vs_add}.
Changing the form of Formula 6 to $\mathcal{L}_{pos}-\mathcal{L}_{neg}$ or \textbf{InfoNCE} leads to artifacts in the subject that disrupt the specified motion trajectory.
In contrast, our formula produces high-quality video results that are semantically consistent with the text prompts.

\subsubsection{Get CA Maps from which Layer.}

The quantitative results are shown in \tabref{tab_layers}.
The CA maps is obtained from the lowest resolution layer by integrating upsampling and downsampling, which produces optimal results across all indicators and significantly outperforms other combinations. 
In terms of numeracy correctness, the approach improves $0.224$, $0.055$, and $0.12$ compared to results obtained using only downsampling layers, only upsampling layers, or a combination of upsampling, downsampling, and intermediate layers.

\begin{table}[!t]
\centering
\caption{\textbf{Ablation study of CA Map's layers.}
The best value is highlighted in \colorbox{pearDark!20}{blue}.
}
\resizebox{\linewidth}{!}{%
\begin{tabular}{cccccc}
\hline
\multicolumn{1}{c}{} & \multicolumn{2}{c}{\textbf{Video Quality}} & \multicolumn{3}{c}{\textbf{Semantic Alignment}} \\ 
\cmidrule(r){2-3} \cmidrule(r){4-6}
\multicolumn{1}{c}{\multirow{-2}{*}{\textbf{Method/Metrics}}} & \textbf{Pick Score} \textbf{($\uparrow$)} & \textbf{CLIP-I} \textbf{($\uparrow$)} & \textbf{CLIP-T} \textbf{($\uparrow$)} & \textbf{Numeracy} \textbf{($\uparrow$)} & \textbf{Action Binding} \textbf{($\uparrow$)} \\ 
\midrule
only down & 20.17 & \colorbox{pearDark!20}{0.94} & 25.64 & 0.647 & 0.665 \\
only up & 20.36 & \colorbox{pearDark!20}{0.94} & 25.87 & 0.816 & 0.748 \\
up + mid + down  & 20.61 & \colorbox{pearDark!20}{0.94} & 26.13 & 0.751 & 0.727 \\
\textbf{Ours} & \colorbox{pearDark!20}{20.69} & \colorbox{pearDark!20}{0.94} & \colorbox{pearDark!20}{27.76} & \colorbox{pearDark!20}{0.871} & \colorbox{pearDark!20}{0.795}  \\
\bottomrule
\end{tabular}%
}
\label{tab:tab_layers}
\end{table}

\subsection{The Analysis of Computational Overhead}
\label{sec:supp_cost}

Our approach calculates the gradient of the proposed attention-based constraint and employs it to update the noisy latent to improve the semantic alignment between multiple subjects, their motions, and textual prompts in the pre-trained T2V model. 
Compared to the pre-trained T2V model, this leads to an increase in VRAM and inference time.
As shown in \tabref{tab_cost}, our method's inference time is approximately three times that of the baseline model. 
Despite the increased computational overhead, our method achieves superior performance, as demonstrated by the qualitative and quantitative results in Fig. 5 and Table I in the main text.

\subsection{Additional Results}
\label{sec:supp_addr}

\begin{figure*}[h]
    \centering
    \includegraphics[width=0.95\linewidth]{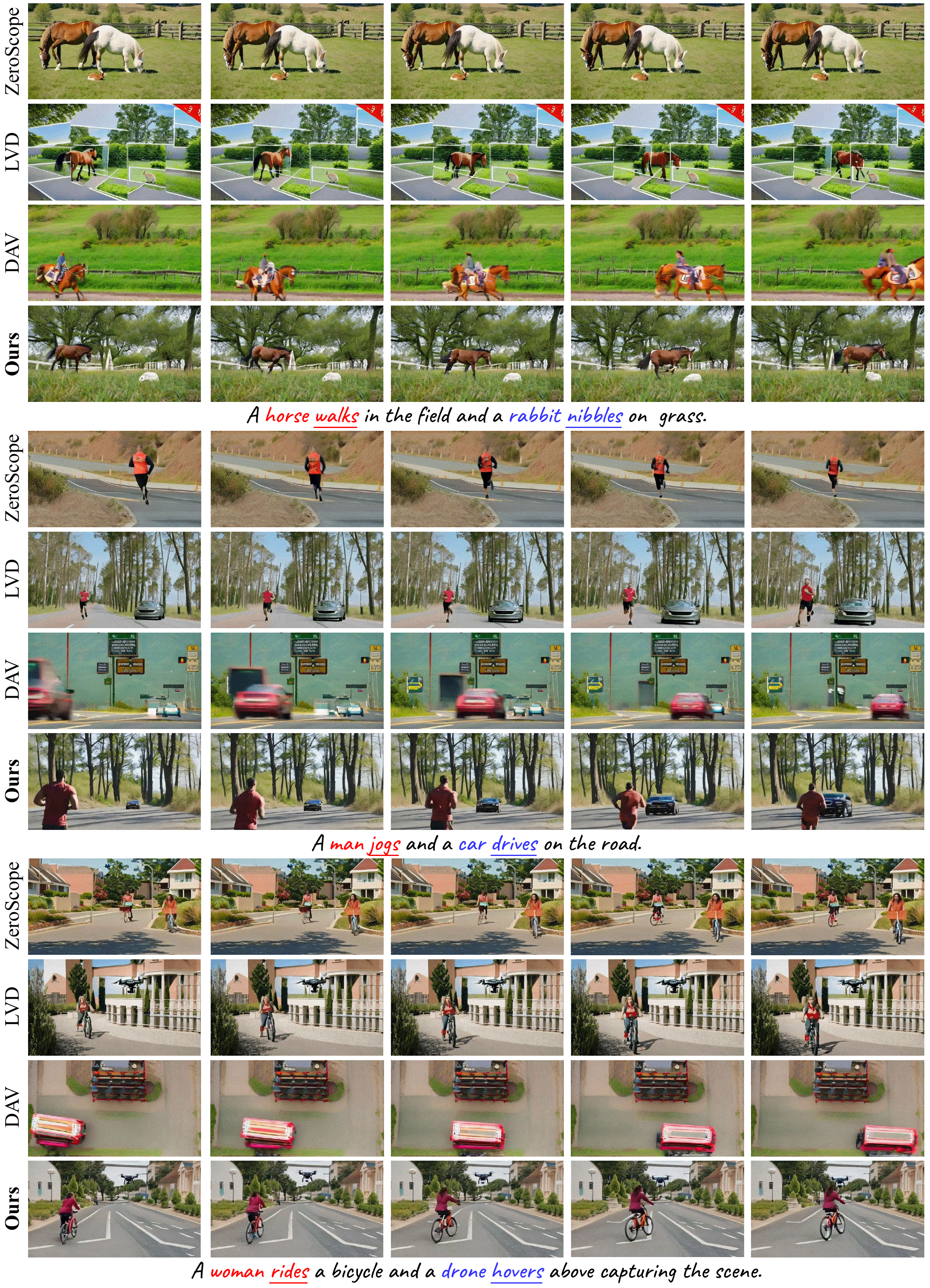}   
    \vspace{-4 mm}
    \caption{Qualitative comparison with ZeroScope~\cite{wang2023modelscope} on Action Binding Benchmark.}
    \vspace{-3 mm}
    \label{fig:demo_zeroscope_ac_1}
\end{figure*}

\begin{figure*}[h]
    \centering
    \includegraphics[width=0.95\linewidth]{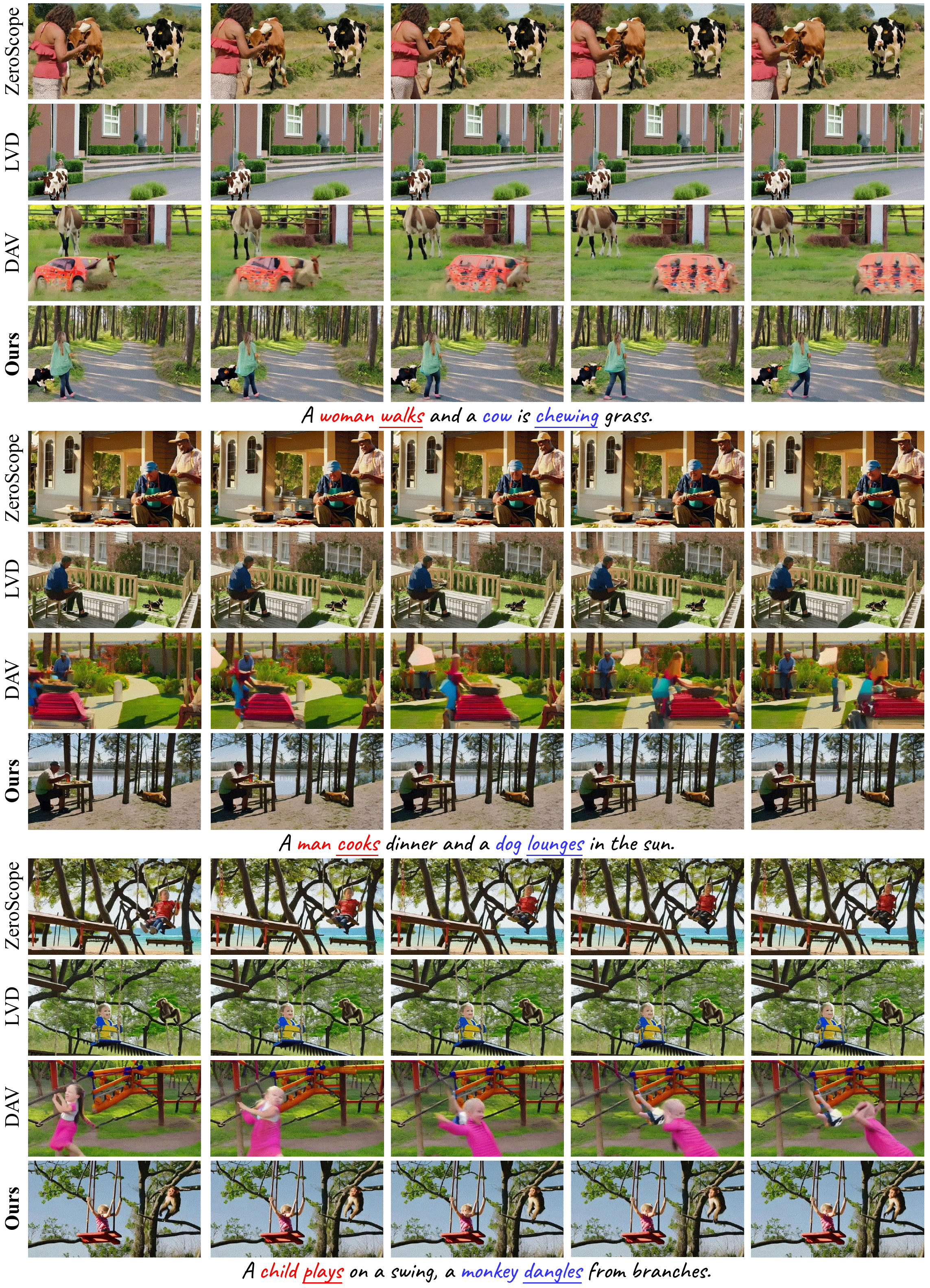}   
    \vspace{-4 mm}
    \caption{Qualitative comparison with ZeroScope~\cite{wang2023modelscope} on Action Binding Benchmark.}
    \vspace{-3 mm}
    \label{fig:demo_zeroscope_ac_2}
\end{figure*}

\begin{figure*}[h]
    \centering
    \includegraphics[width=0.95\linewidth]{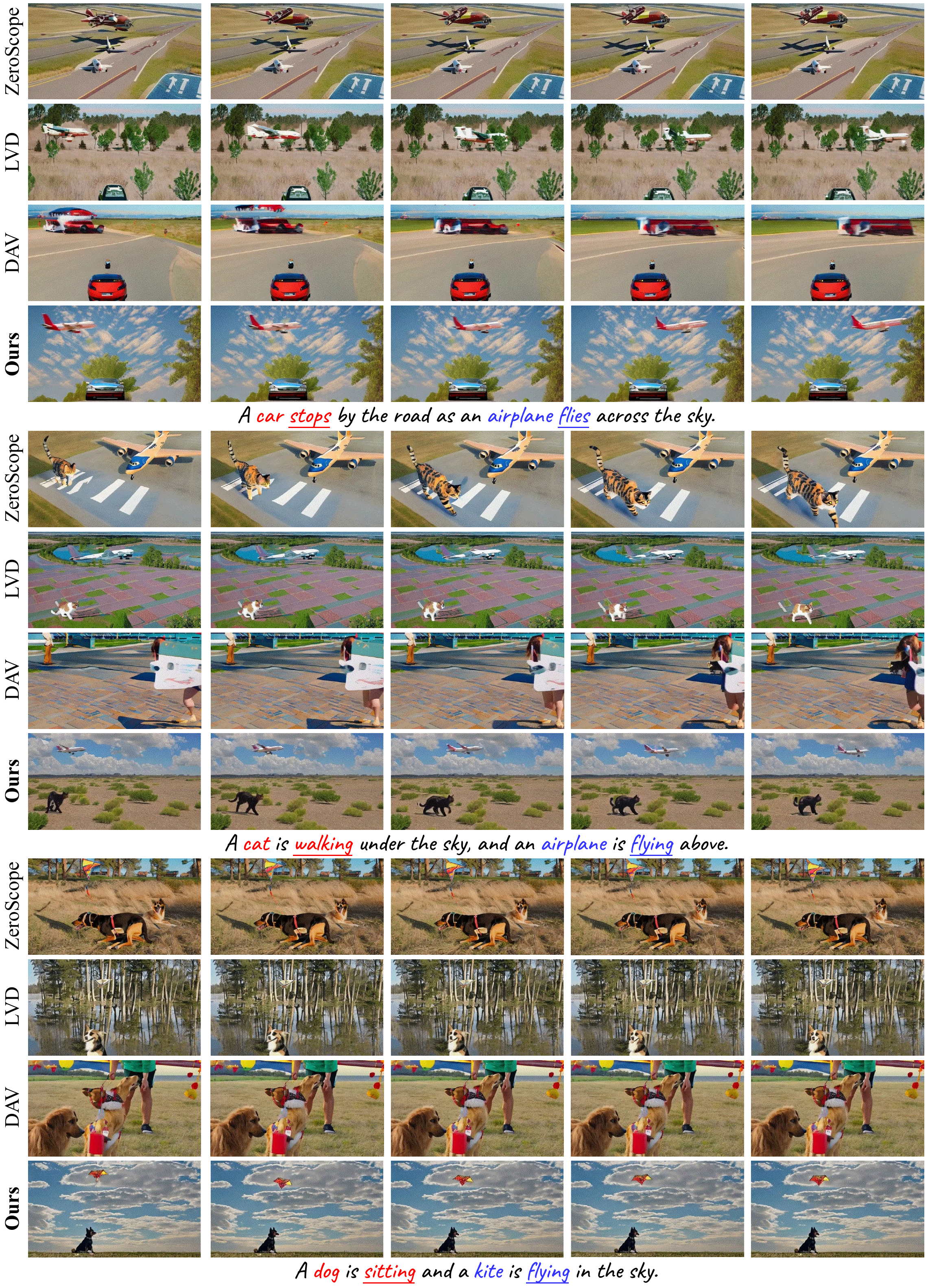}   
    \vspace{-4 mm}
    \caption{Qualitative comparison with ZeroScope~\cite{wang2023modelscope} on LLM-Generated Benchmark.}
    \vspace{-3 mm}
    \label{fig:demo_zeroscope_LLM_1}
\end{figure*}

\begin{figure*}[h]
    \centering
    \includegraphics[width=0.95\linewidth]{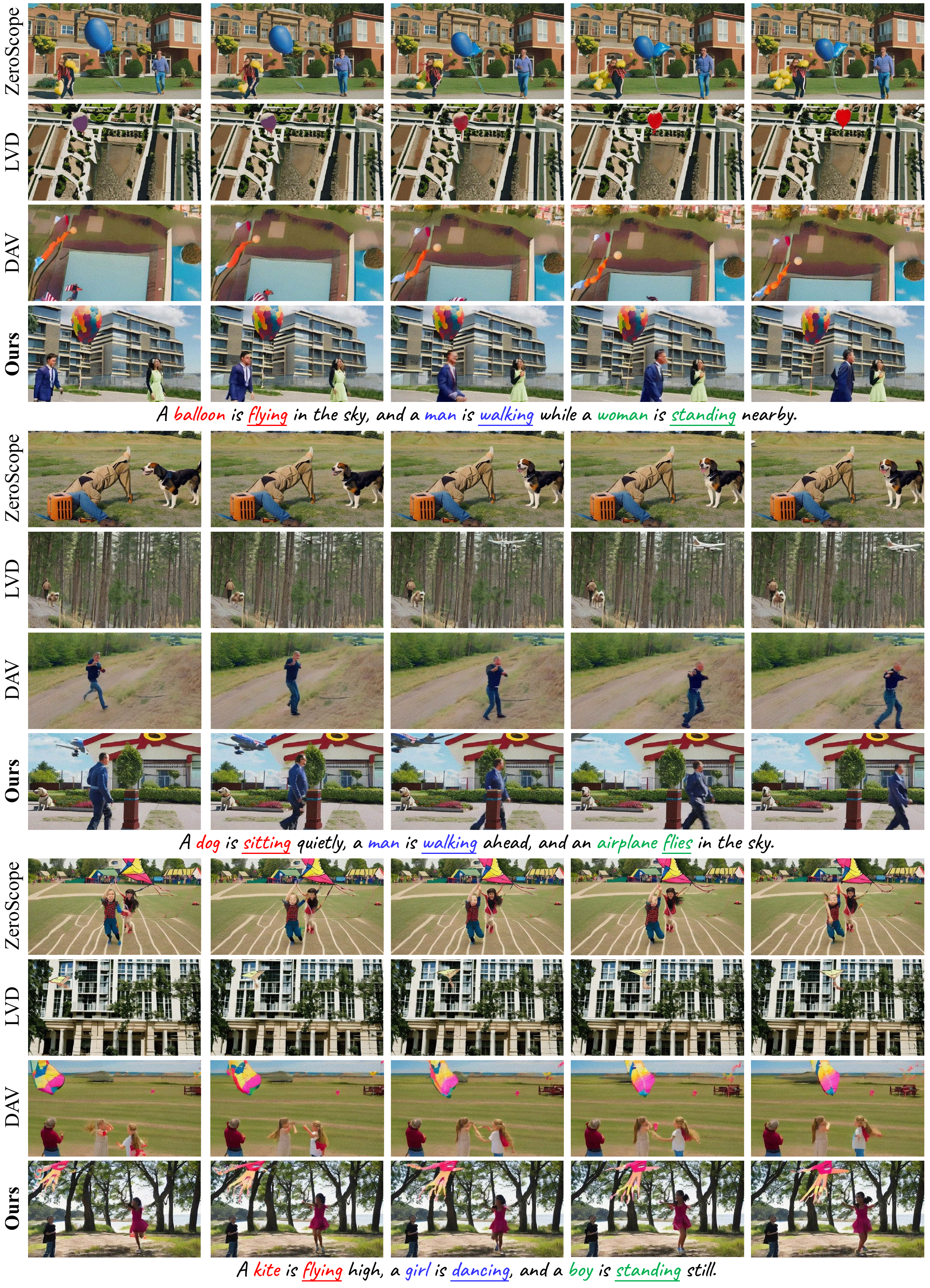}   
    \vspace{-4 mm}
    \caption{Qualitative comparison with ZeroScope~\cite{wang2023modelscope} on LLM-Generated Benchmark.}
    \vspace{-3 mm}
    \label{fig:demo_zeroscope_LLM_2}
\end{figure*}

\begin{figure*}[h]
    \centering
    \includegraphics[width=0.95\linewidth]{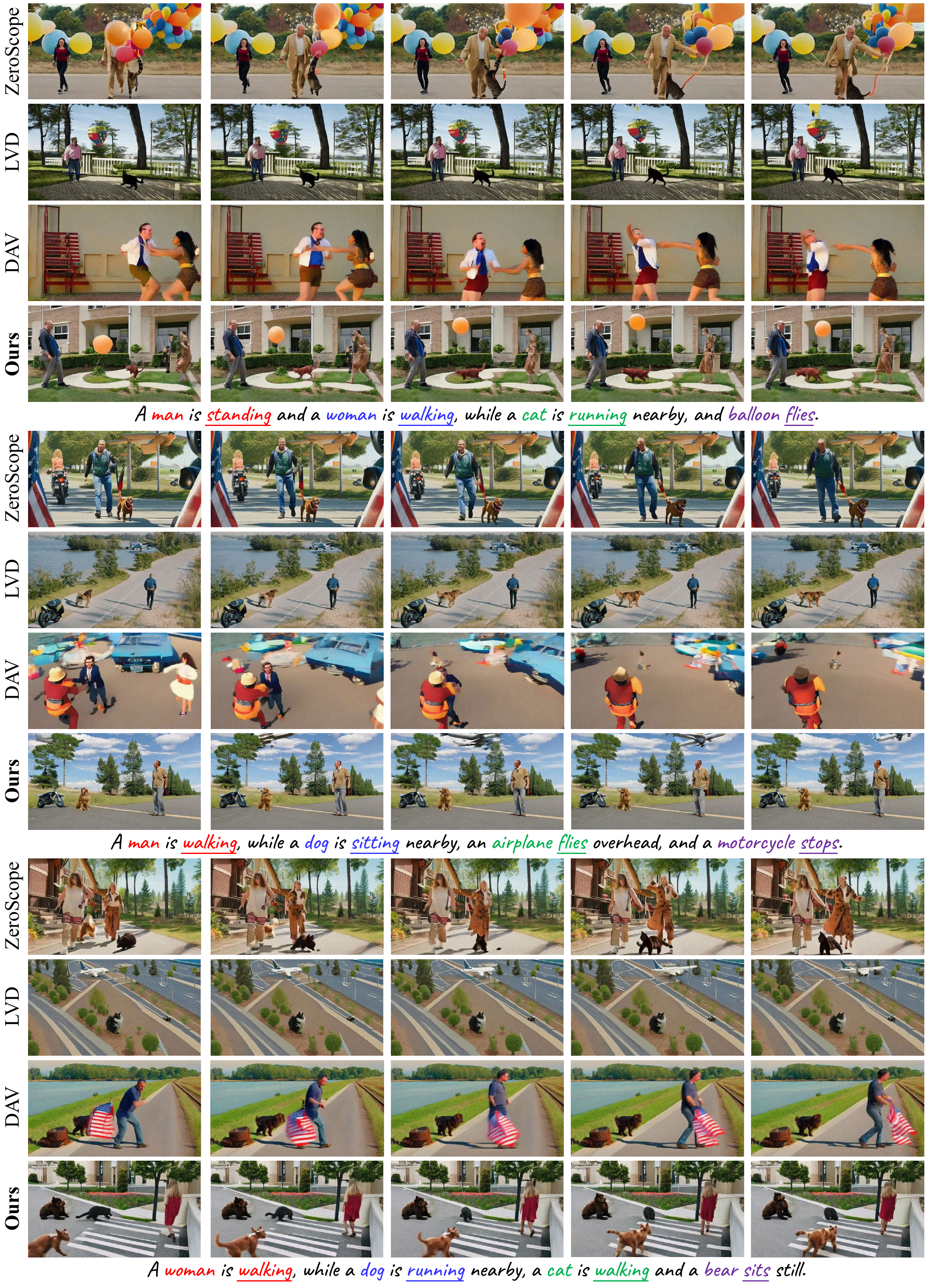}   
    \vspace{-4 mm}
    \caption{Qualitative comparison with ZeroScope~\cite{wang2023modelscope} on LLM-Generated Benchmark.}
    \vspace{-3 mm}
    \label{fig:demo_zeroscope_LLM_3}
\end{figure*}

\begin{figure*}[h]
    \centering
    \includegraphics[width=0.95\linewidth]{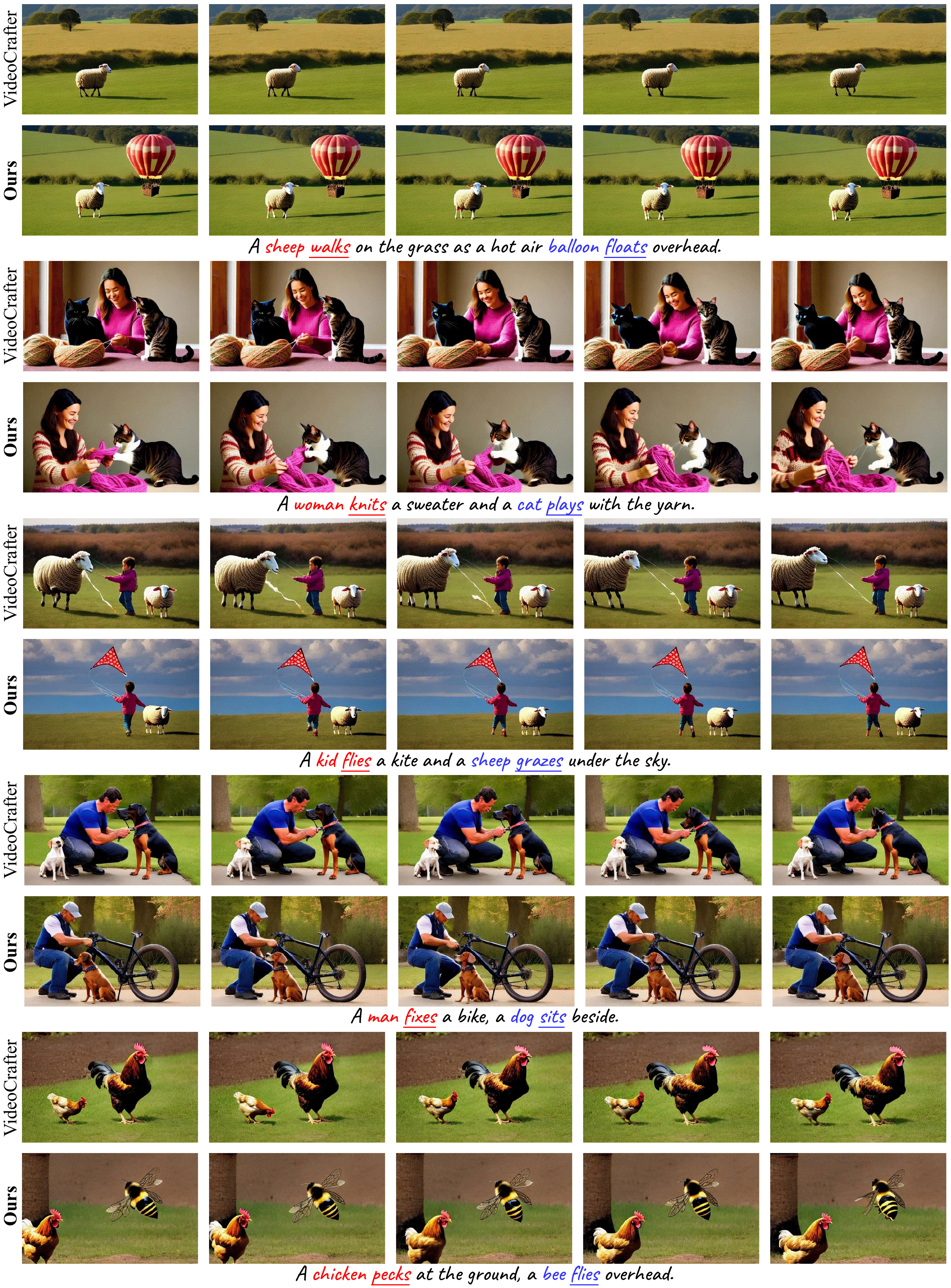}   
    \vspace{-4 mm}
    \caption{Qualitative comparison with VideoCrafter2~\cite{chen2024videocrafter2} on LLM-Generated Benchmark.}
    \vspace{-3 mm}
    \label{fig:demo_VideoCrafter_ac_1}
\end{figure*}

\figref{demo_zeroscope_ac_1} to \figref{demo_zeroscope_LLM_3} demonstrate additional qualitative comparison results of our method with ZeroScope~\cite{wang2023modelscope}, LVD~\cite{lian2023llmgroundedvideo}, and DAV~\cite{yang2024direct} on both benchmarks.
Meanwhile, \figref{demo_VideoCrafter_ac_1} and \figref{demo_VideoCrafter_LLM_2} provide further visual comparisons with VideoCrafter2~\cite{chen2024videocrafter2}.

\begin{table}[!t]
\centering
\caption{\textbf{The computational overhead of our approach.}
}
\resizebox{\linewidth}{!}{%
\begin{tabular}{ccccc}
\hline
Method & resolution & frames & VRAM & inference time  \\
\midrule
ZeroScope & $320 \times 576$ & 16 & 7068MB & 40.52s \\
Ours & $320 \times 576$ & 16 & 30040MB & 122.12s \\
\bottomrule
\end{tabular}%
}
\label{tab:tab_cost}
\end{table}


\begin{figure*}[h]
    \centering
    \includegraphics[width=0.95\linewidth]{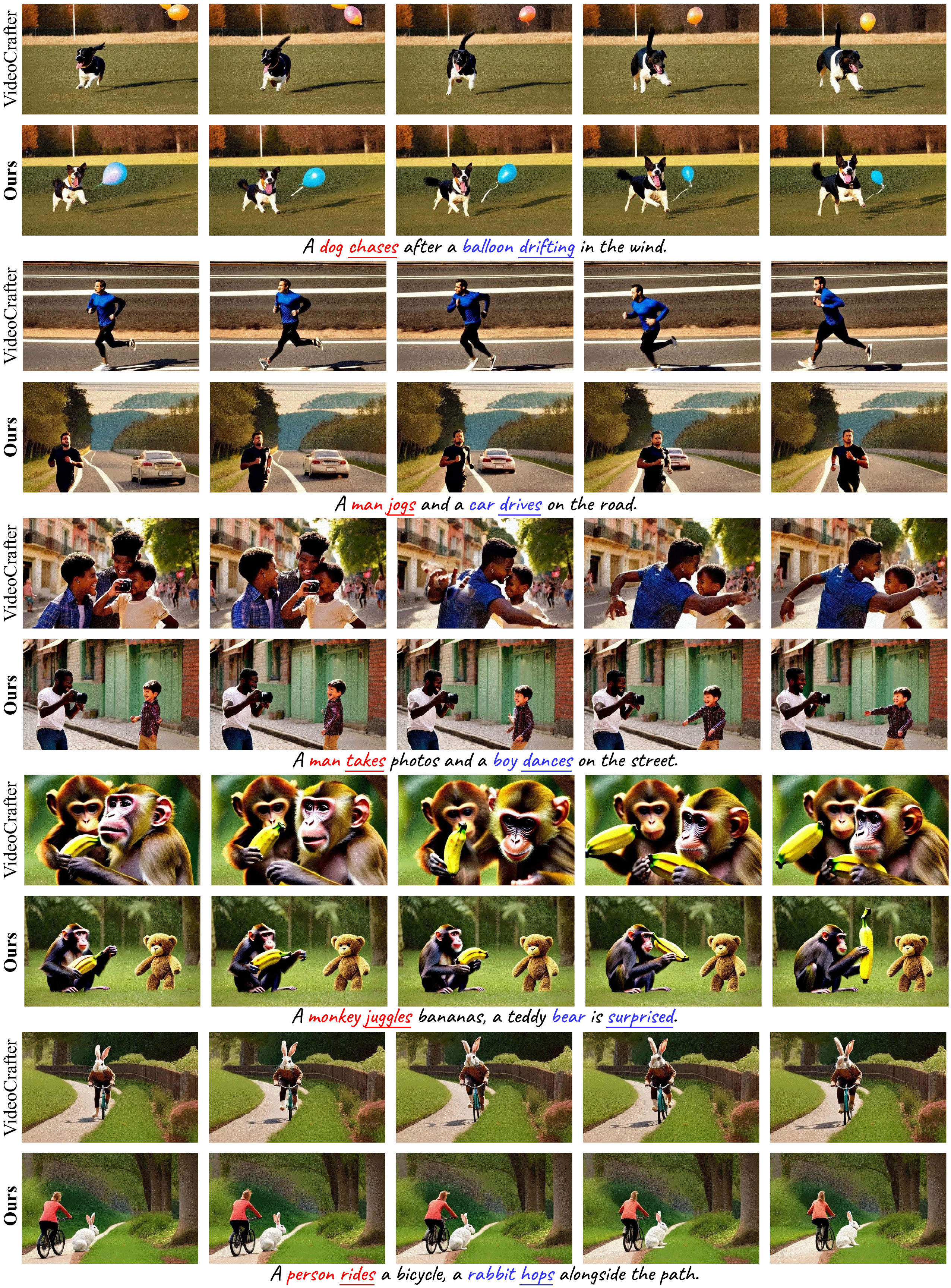}   
    \vspace{-4 mm}
    \caption{Qualitative comparison with VideoCrafter2~\cite{chen2024videocrafter2} on LLM-Generated Benchmark.}
    \vspace{-3 mm}
    \label{fig:demo_VideoCrafter_ac_2}
\end{figure*}
\begin{figure*}[h]
    \centering
    \includegraphics[width=0.95\linewidth]{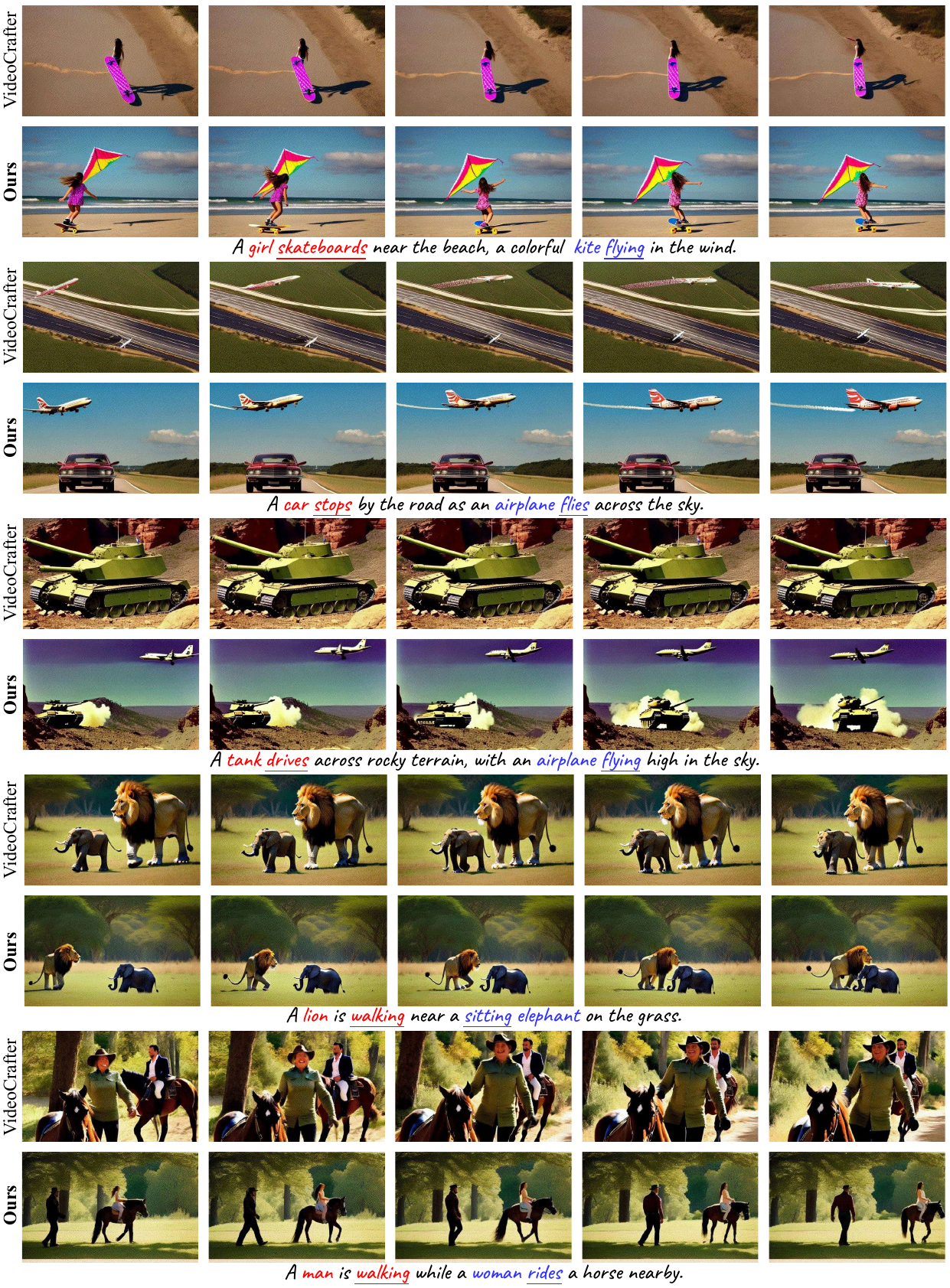}   
    \vspace{-4 mm}
    \caption{Qualitative comparison with VideoCrafter2~\cite{chen2024videocrafter2} on LLM-Generated Benchmark.}
    \vspace{-3 mm}
    \label{fig:demo_VideoCrafter_LLM_1}
\end{figure*}
\begin{figure*}[h]
    \centering
    \includegraphics[width=0.95\linewidth]{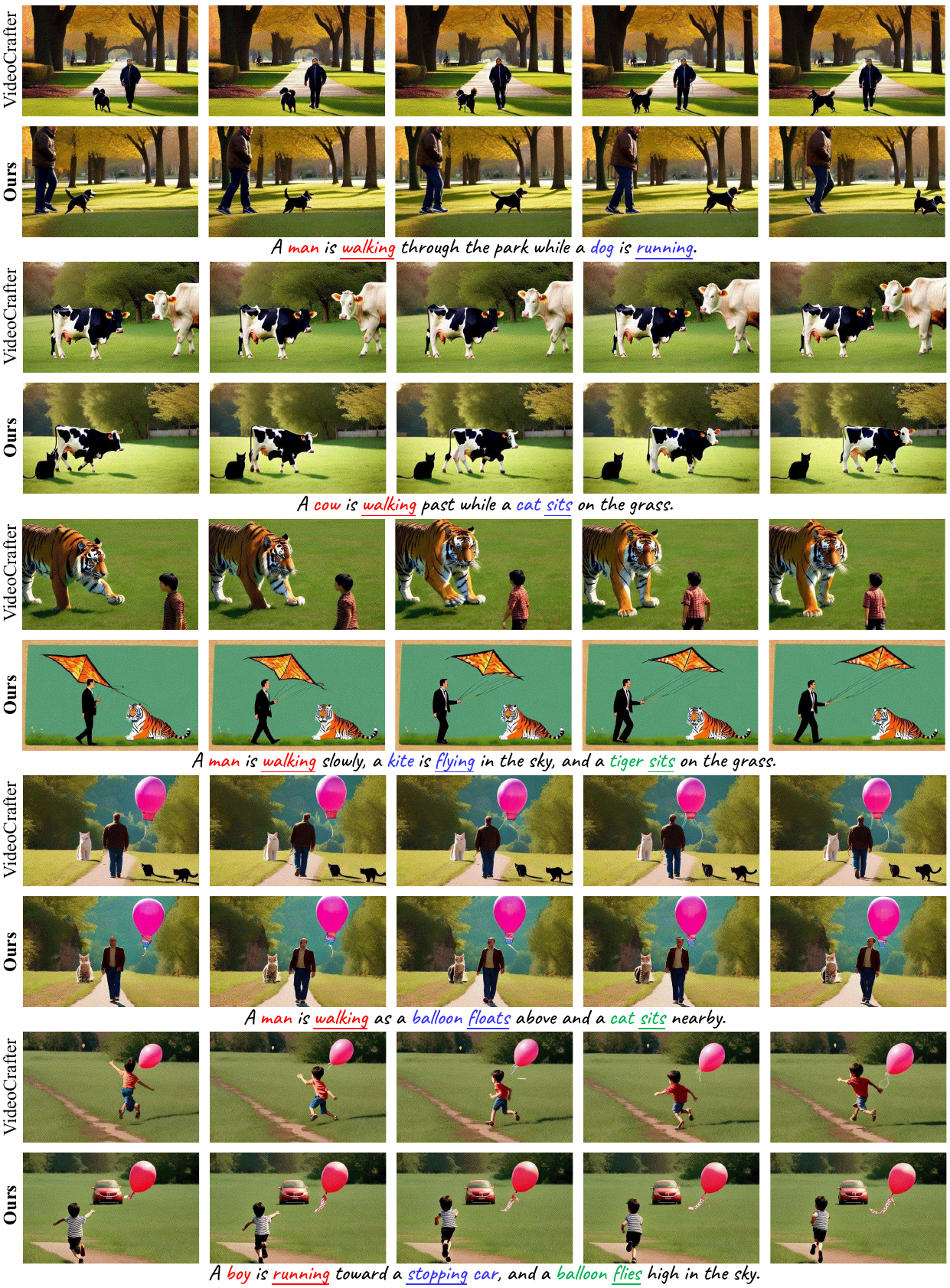}   
    \vspace{-4 mm}
    \caption{Qualitative comparison with VideoCrafter2~\cite{chen2024videocrafter2} on LLM-Generated Benchmark.}
    \vspace{-3 mm}
    \label{fig:demo_VideoCrafter_LLM_2}
\end{figure*}

\end{document}